%% file: main.tex
\documentclass[runningheads]{llncs}

\usepackage{eccv}

\usepackage{eccvabbrv}

\usepackage{graphicx}
\usepackage{booktabs}
\usepackage{algorithm}
\usepackage{algorithmic}
\usepackage[table]{xcolor}
\usepackage{multirow}
\usepackage{graphicx}
\usepackage{tcolorbox}
\tcbuselibrary{listings, breakable}
\usepackage[accsupp]{axessibility}  

\usepackage{hyperref}
\usepackage{tcolorbox}
\usepackage{orcidlink}
\usepackage{wrapfig}
\def\methodName{E-TTS}

\begin{document}

\title{E-TTS: A New Embodied Test-Time Scaling Framework for Robotic Manipulation} 

\titlerunning{\methodName{}}


\author{ Wen Ye\inst{1,2}\textsuperscript{*} \and Peiyan Li\inst{1,2}\textsuperscript{*} \and Tingyu Yuan\inst{2,3} \and Yuan Xu\inst{1,2} \and \\
Xiangnan Wu\inst{1,2} \and Chaoyang Zhao\inst{3} \and Jing Liu\inst{4} \and Nianfeng Liu\inst{4} \and \\
Yan Huang\inst{1,2,4}\textsuperscript{\ensuremath{\dagger}} \and Liang Wang\inst{1,2}\textsuperscript{\ensuremath{\dagger}}}

\begingroup \renewcommand{\thefootnote}{} \footnotetext{ \textsuperscript{*}Equal contribution. \\ \textsuperscript{\ensuremath{\dagger}}Corresponding authors. } \addtocounter{footnote}{-1} \endgroup

\authorrunning{W. Ye et al.}


\institute{
New Laboratory of Pattern Recognition (NLPR), Institute of Automation,\\
Chinese Academy of Sciences, Beijing, China\\
\and
School of Artificial Intelligence, \\ University of Chinese Academy of Sciences, Beijing, China
\and
Foundation Model Research Center, Institute of Automation,\\
Chinese Academy of Sciences, Beijing, China
\qquad \textsuperscript{4}\,FiveAges, Beijing, China \\
\email{yewen2025@ia.ac.cn, peiyan.li@cripac.ia.ac.cn}\\
\email{\{yhuang, wangliang\}@nlpr.ia.ac.cn}
}


\maketitle

\vspace*{-1.5em} 


\begin{abstract}
Recently, a few works have made early attempts to study test-time scaling for embodied tasks. However, two major challenges remain unsolved: 
(1) reasoning can effectively improve the performance of the policy, but its scaling mechanism has seldom been studied; (2) historical information is essential, as embodied tasks are inherently long-horizon and sequential, making sole reliance on current observations for action scaling inadequate due to the lack of historical context utilization.
To address these challenges, we introduce E-TTS, a modular and plug-and-play Embodied Test-Time Scaling framework that unifies reasoning and action scaling for robotic manipulation via history-aware iterative refinement with vision-language verifiers. 
To support joint reasoning-action scaling, E-TTS performs reasoning-action joint sampling and scoring in a pairwise manner. To better utilize historical information, E-TTS uses a history buffer to store historical context, which is then used by reasoning and action verifiers to evaluate the sampled candidates. 
Unlike conventional open-loop TTS methods, E-TTS introduces feedback generation into the sampling process to form a closed-loop iterative refinement mechanism, enhancing both inference efficiency and environmental adaptability. Each component functions as an independent and composable module, allowing flexible and adaptive configuration depending on task requirements.
To evaluate the advantages of our framework, we conduct experiments across 4 different benchmarks, 6 environments, 3 embodiments, and 4 base vision-language-action models. The experimental results demonstrate that, without requiring additional expert data collection or retraining, E-TTS consistently improves performance, achieving up to a 33.14\% increase in simulation and 26.62\% in real-world scenarios. Our project page is \url{https://27yw.github.io/E-TTS-Web/}.
\end{abstract}

\section{Introduction}
\label{sec:intro}

\begin{figure*}[htbp]
  \centering
  \includegraphics[width=\textwidth]{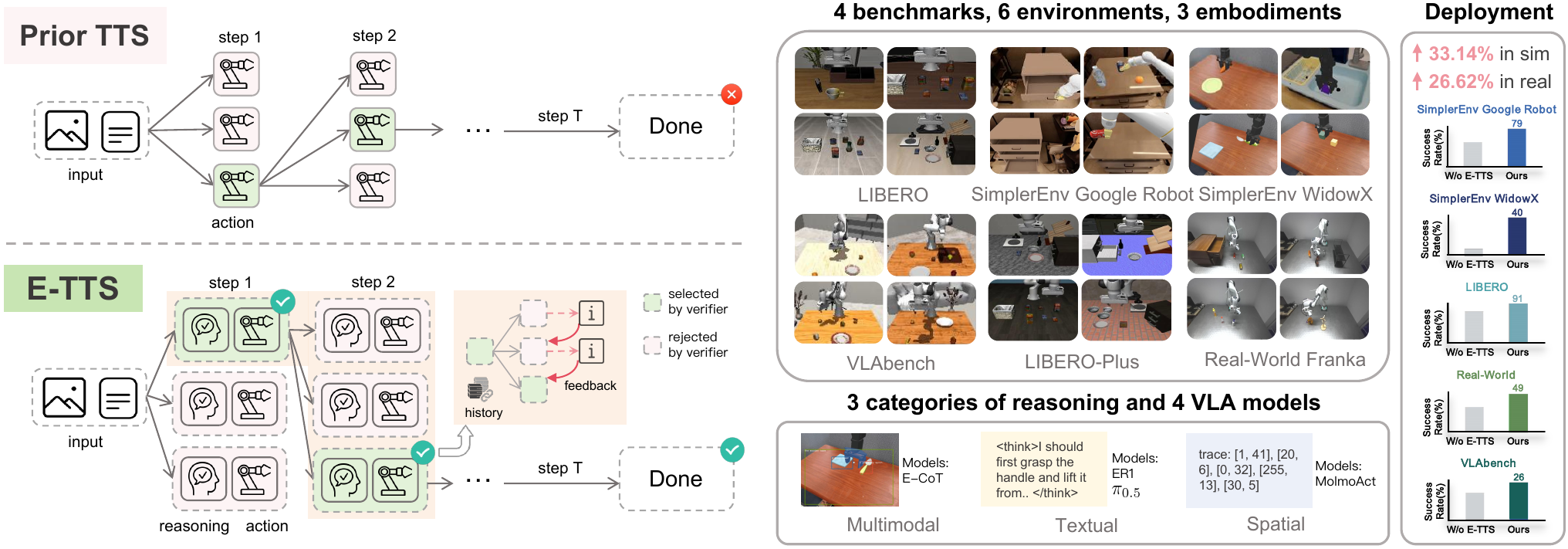}
  \caption{\textbf{Overview.} \methodName{} is an embodied test-time scaling framework that integrates reasoning and action scaling for robotic manipulation through history-aware, closed-loop interactions with vision-language verifiers. When combined with standard VLA models, \methodName{} consistently enhances performance, achieving up to a 33.14\% improvement in simulation and 26.62\% in real-world scenarios.}
\label{fig:teaser}
\end{figure*}

Test-time scaling (TTS) has gained significant attention in fields such as computer vision and natural language processing as it can improve model's performance without requiring additional data or retraining. Such an advantage is even more appealing for the embodied domain, as collecting real-world robotic data is substantially more expensive than obtaining internet vision–language data. Fundamentally, TTS trades additional inference-time computation for higher output quality. 
In the embodied domain, there is a wide range of latency-insensitive tasks (e.g., object rearrangement), which prioritize task success over millisecond-level latency. For these scenarios, TTS offers a promising path toward robust execution. Consequently, a pivotal question arises: how can we apply Test-Time Scaling (TTS) in the embodied domain with an optimal balance between computation and performance?

Recent works~\cite{kwok2025robomonkey,dai2025rover,jang2025verifier} explore test-time scaling (TTS) for embodied systems (see~\cref{fig:teaser}), as shown in the upper part of~\cref{fig:teaser}, yet they overlook two intrinsic challenges of robotic manipulation.
First, it has become popular to incorporate a reasoning component before action prediction~\cite{intelligence2504pi0,zawalski2024robotic,qu2025eo,zhai2025igniting,lee2025molmoact}. However, existing TTS methods focus solely on scaling actions, causing misalignment between high-level planning and execution, and thus suboptimal manipulation performance.
Unlike conventional vision-language tasks, manipulation is a sequential decision-making process. Historical trajectory context is vital for candidate evaluation, while iterative feedback is necessary to refine reasoning and actions. Current TTS approaches lack mechanisms to integrate this history, limiting their effectiveness in complex embodied scenarios.

To effectively address these two challenges, we propose a unified embodied test-time scaling framework (\methodName) that jointly scales reasoning and action through history-aware, closed-loop interaction and refinement with vision–language verifiers (Fig.~\ref{fig:teaser}). 
Specifically, to tackle the first challenge,
\methodName\ performs joint reasoning–action sampling, where the reasoning process and low-level action generation are sampled and selected based on the unified score that is jointly modeled from both dimensions. 
For reasoning selection, we prompt the VLM to perform zero-shot scoring of each reasoning candidate. For action selection, we construct an action preference dataset, based on which a separate VLM is trained as a \textit{verifier} to score candidate actions. 
To address the second challenge, we use a history buffer to store historical information, which is then incorporated into the input context to improve the verifier's ability to model the temporal dependency. Moreover, our framework integrates feedback generation at sampling step. This generated feedback is then incorporated into subsequent sampling rounds, creating an iterative refinement process that enhances both efficiency and adaptability. 
\methodName{} features a highly flexible architecture where every component is optional, independently configurable, and can be combined with others, providing an adaptable framework that accommodates a wide range of tasks and different VLA models without additional training.

To validate its effectiveness, we integrate the proposed \methodName\ with four representative vision-language-action models, all of which take vision-language inputs and output actions, but with different types of intermediate reasoning.
The evaluation experiments are conducted across six environments: SIMPLER WidowX, SIMPLER Google Robot, LIBERO, LIBERO-Plus, VLAbench, and the real world. The average success rates are improved with a maximum gain of 33.14\% and an average gain of 13.52\%.
Additionally, we conduct extensive ablation studies on the key components to identify the sources of improvement and explore the trade-off between reasoning scaling and action scaling, finding that both are important for embodied test-time scaling.

To summarize, our main contributions are threefold:

\begin{itemize}

    \item We introduce a novel plug-and-play \textit{embodied test-time scaling} framework that can be seamlessly integrated with various different vision-language-action models, enhancing their success rates without the need for additional expert data collection or retraining.

    \item We tackle the unique challenges of the embodied domain by jointly scaling reasoning and action through history-aware, iterative refinement with vision–language verifiers, leading to significant performance improvements over conventional test-time scaling methods.  
    
    \item We conduct extensive experiments in both simulated and real-world environments to validate the effectiveness of our framework. The results consistently show that our approach significantly boosts task success rates across a variety of environments and embodiments.
\end{itemize}

\section{Related Work}
\subsection{Vision-Language-Action Models}
With the rapid development of embodied domain~\cite{yuan2025unibyd, li2026multi, li2025gr, cai2026xiaomi, xu2025egodemogen, chen2026verm, chen2025ec, chen2026bridgev2w, yang2026uaor, sun2026revisiting, sun2025collabvla}, vision-Language-Action (VLA) models~\cite{kim2024openvla, intelligence2504pi0, li2025bridgevla} have recently become a popular research direction, aiming to unify visual perception, language understanding, and decision-making within a single framework.
However, simply coupling vision, language and motor signals is insufficient for addressing the reasoning demands in complex embodied environments.
Thus, recent research emphasizes embedding reasoning as explicit action tokens, which is able to externalize internal cognitive processes before generating concrete actions~\cite{zhong2025survey}.
Some reasoning-based VLAs, such as E-CoT~\cite{zawalski2024robotic} and RAD~\cite{clark2025action}, incorporate multimodal structured chain-of-thought (CoT) representations into policy learning by synthesizing large-scale reasoning-action datasets.
Some other works, such as ThinkAct~\cite{huang2025thinkact}, $\pi_{0.5}$~\cite{intelligence2504pi0}, and Embodied-R1~\cite{yuan2025embodied}, generate textual reasoning through reinforcement learning and planning, leveraging multimodal latent planning~\cite{huang2025thinkact}, heterogeneous data co-training~\cite{intelligence2504pi0}, and embodied pointing representations~\cite{yuan2025embodied} to improve long-horizon generalization and spatial reasoning.
%
%
More recently, EMMA-X~\cite{sun2024emma} and MolmoAct~\cite{lee2025molmoact} extend this reasoning-enhanced paradigm toward spatially grounded multimodal action reasoning, in which models predict sub-tasks, scene descriptions, and fine-grained motor trajectories grounded in visual context, demonstrating strong generalization in real-world robotic manipulation.
Despite significant progress, most VLAs rely on scaling dataset and model sizes during training to improve final performance, which limits their adaptability in environments with constraints on data and model size.

\subsection{Test Time Scaling}
Recent LLM progress has shifted focus to test-time scaling, which allocates more inference-phase computation via three primary strategies:
Parallel scaling~\cite{li2025s, brown2024large, renze2024effect, lambert2024tulu} improves reliability by generating multiple candidates and aggregating them through selection mechanisms.
Sequential scaling~\cite{yu2024distilling, wei2022chain, madaan2023self, zhou2022least, chen2023teaching} enables iterative refinement, allowing models to progressively build and revise intermediate reasoning.
Hybrid scaling~\cite{yao2023tree, besta2024graph, wang2024mixture} unifies these strengths, exemplified by Tree-of-Thought and Graph-of-Thought, to provide structured, adaptive exploration and deliberation.
Inspired by these language-centric efforts, we explore test-time scaling for the embodied domain. Our framework jointly scales reasoning and action spaces while leveraging history-aware joint verification and selection to coordinate exploration and refinement during execution.

\subsection{Embodied Test Time Scaling}

Few studies have extended the concept of test-time scaling to the embodied domain, aiming to enhance inference-time performance without retraining the underlying policy.
For example, Hume~\cite{song2025hume} introduces a dual-system framework that augments a Vision-Language-Action (VLA) model with a value head for repeated sampling and cascaded denoising, enabling the model to re-evaluate and refine its action proposals during inference. Similarly, RoboMonkey~\cite{kwok2025robomonkey} explores large-scale reward modeling and synthetic data generation to support the external evaluation of action candidates. RoVer~\cite{dai2025rover} designs a robot process-reward model for VLA systems, enabling them to score and refine candidate actions at inference time. TACO~\cite{yang2025steering} designs a test-time scaling framework using pseudo-count estimation.
However, these methods primarily scale within the action space, neglecting the tight coupling between reasoning and action. In contrast, our \methodName{} jointly scales reasoning and action, significantly improving the success rate. Furthermore, these approaches fail to fully leverage historical context and the feedback, neglecting the sequential decision-making characteristics of the embodied domain. In contrast, we propose a history-aware, framework-level closed-loop framework that effectively addresses these unique challenges.

\begin{figure*}
\centering
\includegraphics[page=1,width=\textwidth,trim=0 0 0 0,clip]{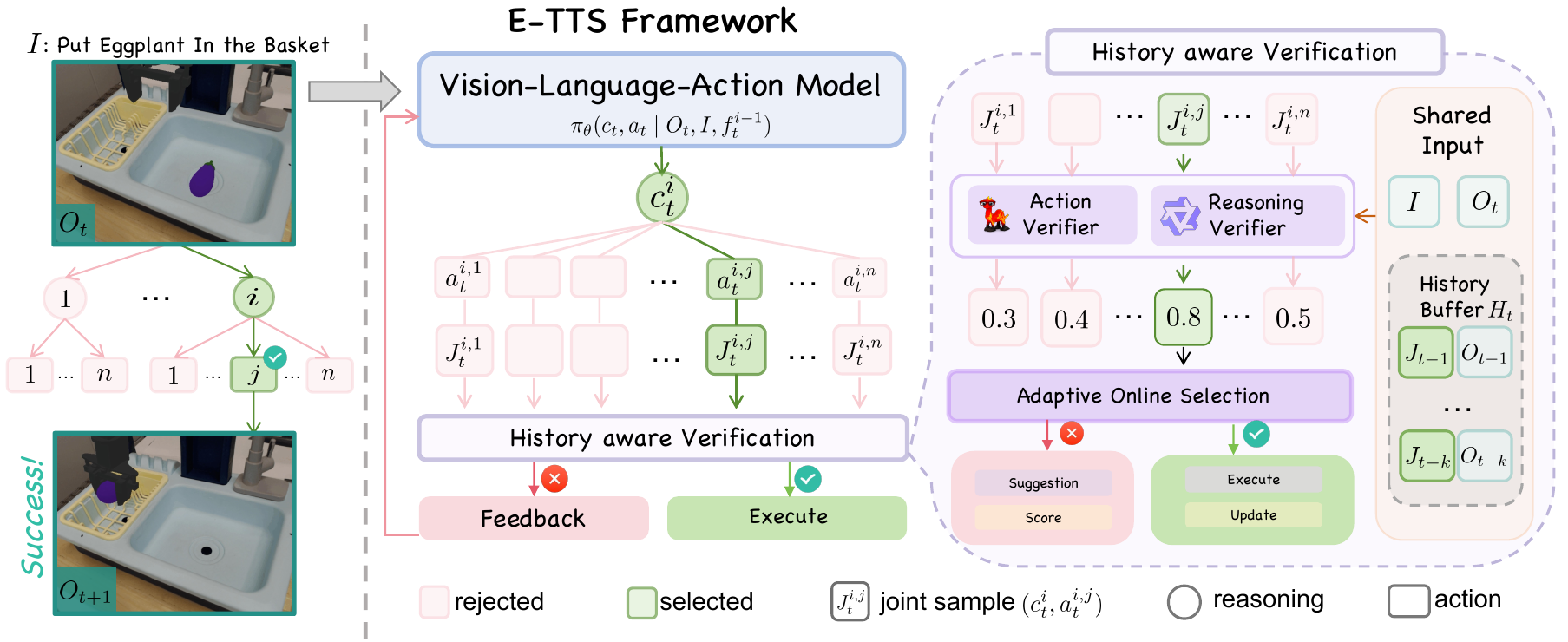}
\captionof{figure}{\textbf{Overview of the proposed \textbf{E-TTS} framework.} 
  At timestep $t$, given an instruction \( I \) and the current observation \( O_t \), the Vision-Language-Action model jointly samples multiple reasoning-action pairs 
  \( J_t^{i,j} = (c_t^i, a_t^{i,j}) \). 
  Each candidate is evaluated by the 
  History-aware Verification module based on history buffer \( H_t \), \( O_t \) and \( I \), which applies both reasoning and action verifiers 
  to assess consistency. 
  Through Adaptive Online Selection, the best pair is executed; otherwise, 
  Feedback-Guided Refinement is triggered to resample improved candidates.}
\label{fig:method}
\end{figure*}
\section{Embodied Test-Time Scaling}
\subsection{Framework Overview}

As shown in Fig.~\ref{fig:method}, given a task instruction \( I \) and the current observation \( O_t \), our method first performs \textbf{Reasoning-Action Joint Sampling (Sec.~\ref{method:joint score})}. This step involves sampling a diverse set of candidate pairs \( J_t^{i,j} = (C_t^i, a_t^{i,j}) \), where \( C_t^i \) represents an intermediate reasoning step and \( a_t^{i,j} \) is the corresponding action.
These joint samples, along with previous observations, are stored in a history buffer denoted as \( H_t \). The buffer is then utilized by our \textbf{History-aware Verification and Selection (Sec.~\ref{sec:verification})} module, which employs a dual-verifier (\textbf{Reasoning (Sec.~\ref{sec:verification_reasoning})} and \textbf{Action (Sec.~\ref{sec:verification_action})}) to assess the candidates’ semantic and physical consistency through a joint scoring strategy.
Next, we apply \textbf{Adaptive Online Selection (Sec.~\ref{sec:verification_adaptive})} to evaluate the suitability of the candidate pair. If the confidence of the selected pair exceeds a predefined threshold, the corresponding action is executed. Otherwise, if no satisfactory candidate is found, the \textbf{Feedback-Guided Iterative Refinement (Sec.~\ref{sec:feedback})} mechanism is triggered, providing corrective feedback to the VLA model. This feedback guides the model to resample more accurate and contextually aligned pairs.

\methodName{} is designed as a plug-and-play framework, and all modules are fully modular and configurable: they can be independently enabled, disabled, or combined depending on the task complexity, allowing \methodName{} to adapt to different tasks and VLA architectures while maintaining a unified framework. The pseudocode for the entire pipeline is presented in Appendix~\ref{sec:Algorithm}.


\subsection{Reasoning-Action Joint Sampling}
\label{method:joint score}

Prior works~\cite{kwok2025robomonkey,dai2025rover,jang2025verifier} primarily focus on scaling the action space, often overlooking the intermediate reasoning process. However, in complex embodied tasks, reasoning outcomes guide action predictions and, consequently, influence the final action results. To address this, \methodName{} introduces a reasoning-action joint scaling strategy that aims to identify the optimal $\langle \text{reasoning}, \text{action} \rangle$ pair rather than selecting a single action.

Specifically, we employ a general robot policy $\pi_{\theta}(c_t, a_t \mid O_t, I)$, where $O_t$ denotes the current observation, $I$ represents the task instruction, $c_t$ is the reasoning result, and $a_t$ is the executed action. At each timestep $t$, we first sample a reasoning $c_{t}^i$ ($i \in [1, M]$), based on which the policy generates $N$ candidate actions $\{ a_t^{i,j} \}_{j=1}^{N}$. We define each $\langle \text{reasoning}, \text{action} \rangle$ pair as a joint unit, and the collection of all pairs corresponding to the same reasoning sample forms a joint sample batch:
\begin{equation}
J_t^{i,j} = (c_t^i, a_t^{i,j}), \, \mathcal{B}_{t}^i = \{ J_t^{i,j} \}_{j=1}^{N}
\end{equation}
where $M$ is the maximum number of reasoning samples at each timestep and $N$ is the number of action samples for each reasoning. 
This joint sampling paradigm generates a structured search space of reasoning-action pairs for the subsequent verification and selection module.

Furthermore, since embodied tasks are inherently sequential decision processes, the final decisions depend not only on the current observation but also on the historical context. To account for this, we store the selected $\langle \text{reasoning}, \text{action} \rangle$ pairs along with past observations in a history buffer, providing historical experience for other modules.

Specifically, at timestep $t$, we maintain a dynamic buffer:
\begin{equation}
\mathcal{H}_t = \{ J_{t-K}, O_{t-K}, \ldots, J_{t-1}, O_{t-1} \}
\end{equation}
which stores the most recent $K$ reasoning–action pairs.
This history buffer operates in a sliding-window manner with a total length of $K$: when a new joint pair $J_{t+1}$ is generated and verified, it is appended to $\mathcal{H}_t$, while the oldest entry $J_{t+1-K}$ is discarded, resulting in $\mathcal{H}_{t+1}$. More details can be found in Appendix~\ref{sec:details_method}.

\subsection{History-aware Verification and Selection}
\label{sec:verification}

Building upon the joint sampling strategy and history buffer, we introduce a dual-verifier mechanism to evaluate the joint samples. 

\subsubsection{Reasoning Verifier ($V_{\text{c}}$)}
\label{sec:verification_reasoning}

We employ a vision–language foundation model ({Qwen2.5-VL-7B~\cite{bai2025qwen2}}) as the reasoning verifier in a zero-shot manner to assess the validity, coherence, and groundedness of candidate reasoning samples. 

Formally, for each $J_t^{i,j}$ in the $\mathcal{B}_{t}^i$, given the history-aware reasoning sequence $\mathcal{H}_t$, the current observation $O_t$, and the task instruction $I$, the reasoning verifier generates a confidence score:
\begin{equation}
S_{\text{c}}^{i,j} = V_{\text{c}}(\mathcal{H}_t, J_t^{i,j}, O_t, I)
\end{equation}
where $S_{\text{c}}$ quantifies the likelihood that the reasoning sample will lead to successful task completion within the current embodied context. 

To address the variety of intermediate reasoning results, we categorize them into three complementary types and design specialized processing strategies:

\textsc{\textbf{Textual Reasoning.}}
This type of reasoning involves purely linguistic content, such as subtasks. Embodied-R1~\cite{yuan2025embodied} is one of the representative manipulation models outputting such intermediate reasoning.  Our reasoning verifier processes these samples through contextual embedding to evaluate logical consistency, causal soundness, and task relevance.

\textsc{\textbf{Multimodal Reasoning.}}
This reasoning integrates both visual perception (e.g., object detection and segmentation results) and textual understanding (e.g., scene descriptions and task decomposition). E-CoT~\cite{zawalski2024robotic} generates this type of reasoning. Such reasoning requires the reasoning verifier to assess cross-modal consistency. 
In our framework, we render the visual perception directly on the image as visual prompts and provide these alongside the original textual understanding to avoid losing critical spatial details.

\textsc{\textbf{Spatial Reasoning.}}
Spatial reasoning typically involves information related to spatial relationships, such as MolmoAct's~\cite{lee2025molmoact} trajectory keypoints. In our framework, we transform these spatial reasoning results into visual prompts and overlay them on the original observations. To help the verifier better understand spatial context, we construct representative examples that show what constitutes a valid reasoning prediction in the current environment. This approach enables $V_{\text{c}}$ to learn to differentiate between well-grounded and inconsistent spatial reasoning, ensuring robust evaluation of spatial intent. Further details on reasoning and our corresponding strategies are provided in Appendix~\ref{sec:re}.

\subsubsection{Efficiency Optimization}
\label{sec:efficiency_Opt}
Introducing reasoning scaling inevitably brings extra computation. To reduce inference latency, we incorporate several system-level optimizations. We deployed the verifier using the vLLM engine and utilize PagedAttention to optimize GPU memory utilization. To further accelerate the process, we adopt a prefix-sharing strategy that caches the KV state of static instructions. Notably, our final approach incurs minimal additional time, with a comprehensive analysis provided in Sec.~\ref{sec:time}.

\subsubsection{Action Verifier ($V_{\text{a}}$)} 
\label{sec:verification_action}
Inspired by ~\cite{kwok2025robomonkey}, our action verifier evaluates the low-level feasibility and task relevance of candidate actions.
The action verifier (LLaVA-7B~\cite{sun2024aligning}) is trained on 90k paired demonstrations that include successful and failed policy rollouts. The data are collected from both SimplerEnv and LIBERO by sampling paired good and bad actions. The quality labels are assigned by comparing the sampled actions with ground-truth actions using an MSE-based criterion.
It learns through a modified Bradley–Terry~\cite{bradley1952rank} objective that incorporates graded preference levels between action pairs.
Specifically, it minimizes the difference between the ground-truth and predicted preference margins, enhancing sensitivity to varying action quality.
Following RT-2~\cite{zitkovich2023rt}, we discretize each dimension of the robot’s continuous actions into 256 bins during training.

During inference, for action in the pair $J_t^{i, j}$ in a joint sample batch, the verifier $V_{\text{a}}$ predicts a scalar score:
\begin{equation}
S_a^{i,j} = V_{\text{a}}(J_t^{i, j}, O_t, I)
\end{equation}
which reflects the action feasibility under the current embodied context.
More details can be found in Appendix~\ref{sec:act}.

\subsubsection{Adaptive Online Joint Selection} 
\label{sec:verification_adaptive}

To evaluate the $\langle \text{reasoning}, \text{action} \rangle$ pair $J_t^{i,j}$, we combine the normalized scores from both verifiers. The final joint score $S_{t}^{i,j}$ for each pair $(c_{t}^i, a_{t}^{i,j})$ is computed as:
\begin{equation}
S_{t}^{i,j} = S_c^{i,j} \times \hat{S}_a^{i,j}
\end{equation}
where $S_c^{i,j}$ represents the score of the reasoning trace $c_{t}^i$, and $\hat{S}_a^{i,j}$ is the normalized score of the action sample $a_{t}^{i,j}$ for reasoning $c_{t}^i$. This ensures that only $\langle \text{reasoning}, \text{action} \rangle$ pairs with both coherent reasoning and feasible actions receive high joint scores. The mathematical justification is provided in Appendix~\ref{sec:math}.

While this joint scoring mechanism effectively measures the consistency between reasoning and action, greedily searching to select the pair with the highest score can limit the model's exploration of the solution space, potentially leading to a sub-optimal or local optimum solution. To address this issue, we adopt an adaptive online selection strategy based on an $\epsilon$-greedy policy, which balances exploration and exploitation during inference.

At each timestep $t$, a random variable $u \sim \text{Uniform}(0,1)$ determines the selection mode: with probability $\epsilon$, a random reasoning–action pair is sampled from the current batch $\mathcal{B}_t^i$ for exploration; otherwise, the pair with the highest joint score is chosen greedily: $J_t^{\mathcal{B}_i} = J_t^{i,j^*}, \quad j^* = \arg\max_j S_t^{i,j}$, where $S_t^{i,j}$ denotes the joint score of pair $J_t^{i,j}$.

To further avoid low-quality selections, a threshold $\eta$ is applied:
\begin{equation}
J_t^{\mathcal{B}_i} =
\begin{cases}
\operatorname{Random}(\mathcal{B}_t^i), & u \le \epsilon \\
J_t^{i,j^*}, & u > \epsilon \ \text{and}\  S_t^{i,j^*} > \eta \\
\text{next batch } \mathcal{B}_t^{i+1}, & u > \epsilon \ \text{and}\  S_t^{i,j^*} \le \eta
\end{cases}
\end{equation}
This adaptive policy dynamically allocates computation to promising candidates while maintaining the model's ability to explore the solution space. More details can be found in Appendix~\ref{sec:select}.

\subsection{Feedback-Guided Iterative Refinement}
\label{sec:feedback}

To help the model find a solution more quickly and avoid repeatedly visiting suboptimal regions, the rejection of an entire joint batch triggers a feedback-guided iterative refinement mechanism.

We reapply our reasoning verifier to analyze the failure causes for the rejected samples and generate structured textual feedback. Specifically, the model is prompted with a set of guided questions aimed at eliciting targeted critiques.

Given the current observation $O_t$, instruction $I$, and the failed sample $J_t^{i,j}$, the verifier produces a textual suggestion $F_{t}^{i}$ that describes the reasoning flaw and offers actionable corrections. This feedback $F_{t}^{i}$ is then inserted into the original instruction prompt, forming a revised context that conditions the next round of joint batch sampling $\mathcal{B}_{t}^{i+1}$:
\begin{equation}
I_{t}^{i+1} = \operatorname{Concat}(I, F_{t}^{i})
\end{equation}
where $I_{t}^{i+1}$ denotes the refined instruction that conditions the base policy.

This closed-loop refinement allows the model to learn from its own errors, progressively improving reasoning coherence and action feasibility until a high-score solution is obtained or a maximum batch limit $M$ is reached.

We provide additional details, including prompts (\ref{sec:prompt}), refinement strategies (\ref{sec:details_method}), and examples of iterative refinement (\ref{sec:example_feedback}) and detailed suggestions (\ref{sec:example_Supervisor}).

\section{Experiments}
We conduct extensive experiments to evaluate the effectiveness of \methodName. 
Our experiments are designed to address the following research questions:

\textbf{Q1:} Is \methodName{} general enough to enhance the performance of various manipulation methods that rely on different intermediate reasoning results?

\textbf{Q2:} Are specific designs for embodied scenarios, such as jointly scaling reasoning and action and incorporating feedback and history, truly beneficial?

\textbf{Q3:} What is the efficiency–performance trade-off of E-TTS, and can it achieve substantial gains without highly excessive latency?

\textbf{Q4:} What is the optimal trade-off between reasoning and action scaling within a fixed computational budget?

\textbf{Q5:} Does \methodName{} maintain strong performance in real-world environments?

\subsection{Experiments on Different Types of Manipulation Methods}
\label{exp:different_methods}
We evaluate \methodName{} on four representative vision-language-action models: E-CoT~\cite{zawalski2024robotic}, MolmoAct~\cite{lee2025molmoact}, $\pi_{0.5}$~\cite{intelligence2504pi0} and Embodied-R1~\cite{yuan2025embodied}. 
While all map multimodal inputs to actions, their intermediate reasoning differs.
Specifically, E-CoT generates reasoning outputs including object bounding boxes, gripper positions, subtasks, and motion primitives; 
MolmoAct produces depth predictions and visual traces; 
$\pi_{0.5}$ and Embodied-R1 generates task plans. 
More details about these three models are provided in Appendix~\ref{sec:base_model}. 

We further compare against RoboMonkey~\cite{kwok2025robomonkey} and a naive test-time scaling baseline. 
This naive test-time baseline also scales the reasoning and action, verifying them through the same vision-language foundation model as ours. However, it lacks other embodiment-related adaptations, such as closed-loop feedback or history-based inputs. For fair comparison, RoboMonkey is also integrated with E-CoT in SimplerEnv WidowX. RoboMonkey scales only actions with the same setting of \methodName{}. The evaluation interval is 10 steps in the experiments. Detailed hyperparameter settings are presented in Appendix~\ref{sec:impl}.

\input{table/ecot}
\textbf{Experiments on E-CoT.}
We integrate \methodName{} into E-CoT and evaluate it on the SimplerEnv benchmark~\cite{li24simpler}. 
SimplerEnv is a suite of real-to-sim environments designed for evaluating robot policies in simulation. 
It provides a standardized arena for benchmarking the success rates of robot policies developed for private real-world platforms such as \textit{Google Robot} and \textit{WidowX}. 
In this paper, we evaluate E-CoT on the \textit{WidowX} platform, using the BridgeData V2~\cite{walke2023bridgedata}, which contains 60{,}096 trajectories collected from 24 environments. 
Each method is evaluated over three task types with 24 trials per task. 
The experimental results are reported in Tab.~\ref{tab:ecot}. 
We observe that the original E-CoT baseline performs poorly, and it often fails to accurately approach the target object. 
In contrast, when integrated with our \methodName{} pipeline, E-CoT produces more appropriate reasoning outputs, resulting in a substantial improvement in average task success rate from 6.67\% to 39.81\%. Compared to the Robomonkey, our method achieves a significant $50.91\%$  improvement in average success rate, while maintaining a highly competitive inference efficiency with only a marginal $7.8\%$ increase in execution time. More results can be found in Appendix~\ref{sec:ecot}.
\input{table/molmoact}

\textbf{Experiments on MolmoAct.}
We evaluate \methodName{} using MolmoAct on SimplerEnv (\textit{Google Robot}), LIBERO~\cite{liu2023libero}, and LIBERO-Plus~\cite{fei2025libero}. For SimplerEnv, we test across two settings: (1) Visual Matching, mirroring training setups, and (2) Variant Aggregation, featuring randomized visual conditions. We integrate our framework into both the official zero-shot and fine-tuned MolmoAct checkpoints and the overall success rate is the average across all scene variants. As shown in Tab.~\ref{tab:molmoact_simpler}, \methodName{} consistently boosts success rates across all settings, significantly outperforming the naive test-time scaling baseline.


\input{table/libero_molmoact}

In the LIBERO benchmark, a simulated Franka Emika Panda arm performs manipulation tasks conditioned on multimodal demonstrations, which consist of front and wrist camera observations, natural language instructions, and delta end-effector pose actions.
LIBERO provides four task suites—\textit{Spatial}, \textit{Object}, \textit{Goal}, and \textit{Long}—each containing ten distinct tasks. 
During evaluation, we perform 50 rollouts per task. 
Detailed results are reported in Tab.~\ref{tab:libero_side_by_side}. 
When integrated with \methodName{}, MolmoAct achieves the highest performance across all four task suites, further demonstrating the effectiveness of our approach. More results can be found in Appendix~\ref{sec:molmoact}.

We further evaluate generalization on LIBERO-plus~\cite{fei2025libero}, which introduces 10,030 tasks across seven perturbation dimensions, including viewpoint changes, object layout shifts, robot initialization variations, instruction rewriting, lighting, background textures, and sensor noise. We evaluate every first 200 tasks in each category, 800 tasks in total, as shown in Tab.~\ref{tab:libero_side_by_side}, our method consistently improves MolMoAct across all categories. The improvements under these diverse perturbations demonstrate enhanced robustness and stronger generalization beyond the standard LIBERO setting.

\input{table/rebuttal/vlabench}

\textbf{Experiments on $\pi_{0.5}$.}
We further integrate our framework with $\pi_{0.5}$, a flow-matching-based VLA model with intermediate subtask prediction.
It is evaluated on VLABench~\cite{zhang2025vlabench}, a benchmark specifically designed to assess long-horizon reasoning, world knowledge transfer, semantic instruction understanding, and physical law awareness. It contains composite tasks requiring multi-step planning and implicit intention understanding. As shown in Tab.~\ref{tab:vlabench_results_horizontal}, integrating our method consistently improves $\pi_{0.5}$~\cite{intelligence2504pi0} across most tracks and representative tasks (Details can be found in Appendix~\ref{appendix:pi05}), with average success rates increasing on all reported tracks. Notably, gains are more pronounced on tasks such as \textit{select\_poker}(from 0.46 to 0.63) and \textit{add\_condiment}(from 0.10 to 0.16), which require multi-step spatial reasoning and semantic grounding. These results indicate that under complex and compositional task settings, it enhances decision consistency across steps and improves long-horizon success rates with \methodName{}. The Avg. is of all 12 tasks and full results can be found in Appendix~\ref{appendix:pi05}. Moreover, we evaluate \methodName{} with TACO~\cite{yang2025steering}, which can be found in Appendix~\ref{appendix:pi05}.

\textbf{Experiments on Embodied-R1 (ER-1).}
We evaluate ER-1 on the SimplerEnv \textit{WidowX} environment using the same evaluation protocol as E-CoT. 
Due to space constraints, detailed results are provided in Appendix~\ref{sec:er1}. 
We observe that integrating \methodName{} substantially improves the original model’s average success rate from 39.8\% to 44.9\%.

These results address \textbf{Q1}, demonstrating that \methodName{} can effectively enhance the performance of manipulation methods regardless of the type of their intermediate reasoning processes.
\subsection{Ablation Studies}

To validate the effectiveness of our model design and provide insights for the community, we conduct five ablation studies on the SimplerEnv \textit{WidowX} environment using E-CoT as the base manipulation model. 
The results are in Tab.~\ref{tab:ecot}.

\textbf{w/o feedback} removes feedback from the scaling step, leading to a significant performance drop and demonstrating its importance in the scaling process.

\input{table/rebuttal/Q2}

\textbf{w/o action scaling} and \textbf{w/o reasoning scaling} apply only one scaling component. Using either alone substantially degrades performance, indicating that both are necessary for embodied tasks.

\textbf{w/o joint scoring} evaluates reasoning and action separately. We first select the highest-scoring reasoning, then sample and score actions conditioned on it. This strategy performs poorly, as high-quality reasoning does not guarantee high-quality actions. Effective evaluation requires joint scoring (Sec.~\ref{method:joint score}).

\textbf{w/o history buffer} removes historical inputs during evaluation, reducing success rate, demonstrating the importance of trajectory information.

\textbf{w/o $\epsilon$-greedy} disables exploration, resulting in lower performance and highlighting the need to balance exploration and exploitation.

\textbf{Parameter analysis.} We perform a ablation (right part of Tab.~\ref{tab:latency_ablation}) on history length $K$, threshold $\eta$, and $\epsilon$ exploration. Performance shows that increasing $K$ improves success up to $K=10$ but drops for longer histories. Moderate $\eta$ and small $\epsilon$ achieve the best trade-off between exploration diversity and reasoning precision, addressing sampling scale and efficiency–performance trade-offs.



Overall, these ablation results address \textbf{Q2}, confirming that our design choices are well-motivated and may provide useful insights for future research.

\subsection{Trade-off between performance and latency}
\label{sec:time}


Test-time scaling inherently trades inference time for performance. However, our empirical results show that this trade-off is favorable. As detailed in Table \ref{tab:latency_ablation}, an optimal scaling configuration yields a 150\% relative improvement in success rate with only a 46.6\% increase in latency, reflecting high sampling efficiency.
By integrating the system-level optimizations from Section \ref{sec:efficiency_Opt}, we further reduce this overhead. Compared to Robomonkey, our method achieves a 51.5\% relative gain in average success rate with a marginal 7.8\% increase in execution time. These results directly answer \textbf{Q3}, yielding a highly favorable efficiency–performance trade-off. Furthermore, E-TTS is model-agnostic and compatible with various VLA architectures. Its modular design allows for composable, selective deployment, enabling flexible joint scaling across reasoning and action dimensions to match task complexity while effectively controlling computational overhead and inference latency.

\subsection{Trade-off Between Action Scaling and Reasoning Scaling}
\begin{wrapfigure}{r}{0.5\textwidth} 
  \centering
  \includegraphics[width=\linewidth]{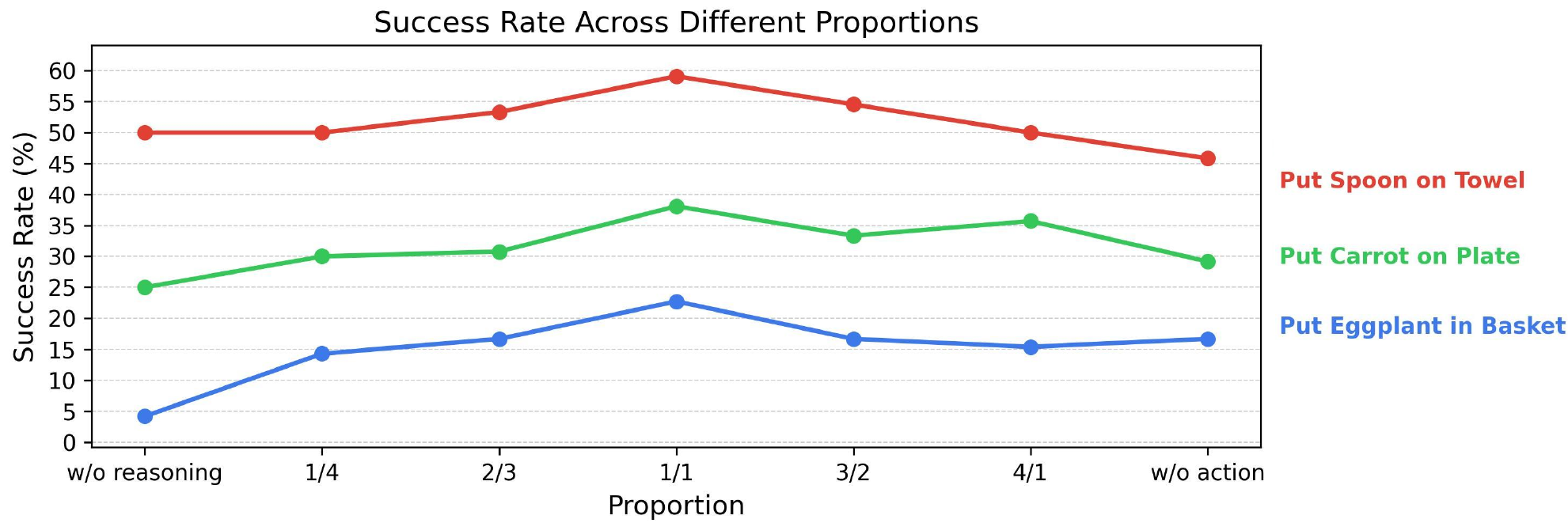} 
  \caption{\textbf{Success rate across different proportions}. The `1/1' proportion achieves peak performance across all tasks.}
  \label{fig:success-rate-ablations}
\end{wrapfigure}
One of the main contributions of this paper is the proposal to scale both reasoning and action. 
A natural question that arises is how to allocate computational resources when they are limited. 
We test five distribution ratios using the E-CoT setup from Sec.~\ref{exp:different_methods}, with a 60-step sampling limit. Results in Fig.~\ref{fig:success-rate-ablations} show that over-prioritizing either component yields suboptimal performance. The peak performance at an approximately 1:1 ratio (\textbf{Q4}) underscores that both reasoning and action scaling are indispensable for effective embodied test-time scaling.


\subsection{Real-robot Experiments}

\begin{figure*}[htbp]
\centering
\includegraphics[page=1,width=0.9\linewidth,trim=0 0 0 0,clip]{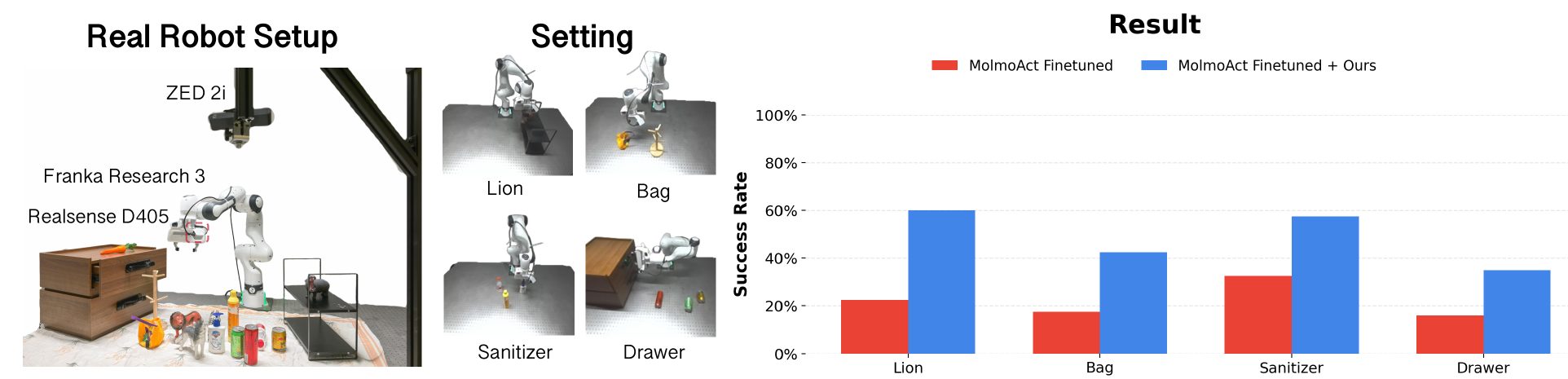}
\captionof{figure}{\textbf{Real-Robot Experiments.}
Overview of the real-world robotic setup and evaluation results. Left: Hardware configuration including ZED 2i camera, Franka Research 3 arm, and RealSense D405. Middle: Representative manipulation tasks. Right: Success rate comparison, where our method (blue) significantly outperforms the MolmoAct-Finetuned baseline, achieving an average improvement of 26.62\%.}
\label{fig:real_world}
\end{figure*}

\textbf{Setup.}
To validate the effectiveness of our framework in real-world settings, we also conduct real-robot experiments. 
Our setup features a Franka Research 3 arm with a parallel-jaw gripper, a static ZED 2i depth camera for global views, and a wrist-mounted Realsense D405 for local observations (Fig.~\ref{fig:real_world}). We evaluate \methodName{} using a finetuned MolmoAct across four tasks, ranging from simple pick-and-place to complex, long-horizon sequences (e.g., placing a coke in a drawer by opening/closing it). These tasks involve both rigid and deformable objects (e.g., a cloth bag). We collected 100 teleoperated expert demonstrations per task via SpaceMouse teleoperation. Each baseline is evaluated over 40 trials per task (160 trials total), with initial object configurations manually standardized for consistency.


\textbf{Results.}
Real-robot results (Fig.~\ref{fig:real_world}) show that \methodName{} boosts the original model's average success rate from 22.13\% to 48.75\%, achieving great performance across all tasks and addressing \textbf{Q5}. Notably, we observed emergent self-correction behaviors: for instance, after a missed grasp, the robot may autonomously re-attempt the action, a capability absent in training. This rollout is visualized in Appendix~\ref{appendix:real_world}. We hypothesize that this stems from the vision-language foundation model’s extensive pretrained knowledge, enabling effective reasoning in out-of-distribution states. More details about task and results are in Appendix~\ref{appendix:real_world}.

\section{Conclusion and Future Work}
We propose an embodied test-time scaling framework that addresses the unique challenges of embodied tasks. It can be integrated with various manipulation methods to enhance performance without requiring additional expert teleoperation data. Through experiments across different environments and embodiments, we validate our method's effectiveness. Though our integrated efficiency optimizations mitigate the latency, future research will explore more acceleration techniques.

\section*{Acknowledgements}
This work was jointly supported by National Natural Science Foundation of China (62322607, 62236010 and 62276261), Beijing Natural Science Foundation (L252033).
%
%
\bibliographystyle{splncs04}
\bibliography{main}

\clearpage
\appendix
\setcounter{page}{1}

\section{Method Details}

\subsection{More Details of E-TTS}
\label{sec:details_method}
Unlike prior work such as RoboMonkey~\cite{kwok2025robomonkey}, which scales only along the action dimension, our framework, \methodName, performs joint reasoning–action scaling within a history-aware, feedback-guided inference loop, enabling coherent coordination between deliberation and control.
We implement this through a joint sampling and verification strategy.

\paragraph{Joint Sampling}
Given a task instruction and the current observation, the VLA model iteratively generates reasoning traces, each providing a high-level plan.
For each reasoning trace, a series of candidate actions is sampled to encourage behavioral diversity while maintaining context relevance.

\paragraph{Joint Verification}
The joint quality of a reasoning–action pair is computed by multiplying the reasoning and action scores.
For the first batch (one reasoning trace with its candidate actions), a random variable ($u \in [0,1]$) is drawn. 
If ($u < \epsilon$), a sample is randomly selected to encourage exploration; otherwise, the highest-scoring sample represents the batch. This sample is compared against a threshold: if it meets the criterion, the action is executed, and the reasoning–action pair along with the observation is updated to the history buffer.
If not, feedback is generated, the current timestep remains unchanged, and the next batch is sampled and verified conditioned on both the feedback and the updated history.

Due to the sequential nature of embodied tasks, which often involve long-horizon action sequences, we design a history buffer and a feedback-guided iterative refinement mechanism. 

\paragraph{History Buffer}
The history buffer stores past reasoning–action pairs and observations, conditioning both sampling and verification to ensure temporal consistency across long-horizon tasks. It is dynamically maintained and incorporated during verification, allowing the model to reference past reasoning–action pairs and observations to ensure temporal consistency and informed decision-making.

\paragraph{Feedback for Refinement}
If no candidate reasoning–action pair in a batch meets the predefined confidence threshold, a feedback-guided refinement is triggered. In this process, the feedback advisor generates structured feedback, providing explicit guidance to correct misalignments and adjust subsequent sampling distributions. The next iteration then samples a new batch conditioned on both the updated feedback and the current task instruction and observation, enabling progressive improvement in both reasoning quality and action precision.

Through this closed-loop process of joint sampling, history-aware verification, and feedback-guided refinement, our method continuously adapts to the task context, maintaining temporal coherence and achieving robust, high-quality performance across long-horizon, sequential embodied tasks.

\subsection{Algorithm of E-TTS}
\label{sec:Algorithm}

\input{table/algorithm}

\subsection{Mathematical Justification of Joint Scoring}
\label{sec:math}
Since the action $a$ is generated conditioned on the reasoning trace $c$, we cannot assume their conditional independence. 
To rigorously justify the multiplicative form of the joint verification score, we start from the exact probabilistic decomposition of task success.

Let $Y=1$ denote success, $c$ the reasoning trace, and $a$ the corresponding action sequence. 
Our goal is to estimate the posterior probability of success given both reasoning and action:
\begin{equation}
    P(Y=1 \mid c, a)
\end{equation}

By the chain rule of probability and Bayes' theorem, we have
\begin{equation}
\begin{aligned}
    P(Y=1 \mid c, a)
    &= \frac{P(c, a \mid Y=1)\, P(Y=1)}{P(c, a)} \\
    &= \frac{P(c \mid Y=1)\, P(a \mid c, Y=1)\, P(Y=1)}{P(c, a)}
\end{aligned}
\end{equation}

We further expand $P(Y=1 \mid c)$ as
\begin{equation}
    P(Y=1 \mid c) = \frac{P(c \mid Y=1)\, P(Y=1)}{P(c)}
\end{equation}
Substituting this into the previous equation yields the following exact identity:
\begin{equation}
\label{eq:exact}
\begin{aligned}
    P(Y=1 \mid c, a)
    &= \frac{P(Y=1 \mid c)\, P(c)\, P(a \mid c, Y=1)}{P(c, a)} \\
    &= P(Y=1 \mid c)\, \frac{P(a \mid c, Y=1)}{P(a \mid c)}
\end{aligned}
\end{equation}

Equation~\ref{eq:exact} shows that the posterior probability of success can be factorized into two interpretable terms:

\begin{itemize}
    \item $P(Y=1 \mid c)$: the probability that the reasoning $c$ leads to success (reflecting high-level plan quality);
    \item $\dfrac{P(a \mid c, Y=1)}{P(a \mid c)}$: a likelihood ratio that measures how much more likely the action $a$ is under successful executions to its general likelihood given $c$. It measures the compatibility of the action with success under the given reasoning.
\end{itemize}

In practice, within a batch of samples conditioned on the same reasoning $c$, the marginal term $P(a \mid c)$ can often be treated as approximately constant, i.e.,
\begin{equation}
    P(a \mid c) \approx C_c
\end{equation}
where $C_c$ is a normalization constant shared across candidates.  
Hence, we obtain the proportional relation
\begin{equation}
\label{eq:approx}
    P(Y=1 \mid c, a) \propto P(Y=1 \mid c)\, P(a \mid c, Y=1)
\end{equation}

Consequently,
\begin{equation}
    S_c \propto P(Y=1 \mid c), \quad S_a \propto P(a \mid c, Y=1)
\end{equation}
then the joint verification score can be expressed as
\begin{equation}
    S_t \propto S_c \times S_a
\end{equation}

Equation~\ref{eq:approx} corresponds to a \textit{joint success likelihood}
Intuitively, a high $S_c$ ensures that the reasoning is coherent and goal-directed, while a high $S_a$ ensures that the chosen action is consistent with successful outcomes under that reasoning.
Their product thus reflects both deliberative soundness and control reliability.

\subsection{Prompts Used in the E-TTS}
\label{sec:prompt}

\paragraph{Prompts for Sampling with Feedback}
We illustrate how feedback-guided refinement is applied in practice. For E-CoT~\cite{zawalski2024robotic}, feedback from previous attempts is used to iteratively adjust reasoning traces and action predictions, improving plan alignment and action precision. The following prompt shows how feedback is incorporated into the next reasoning–action iteration for E-CoT:
\begin{tcolorbox}[breakable, colback=gray!5, colframe=gray!50, boxrule=0.5pt, left=2mm, right=2mm]

\textbf{Task Instruction:}  

What action should the robot take to \{TASK\}  

You are a robotic assistant. Execute the following task based on observations.

\textbf{Previous Attempt:}  

Robot's Reasoning: \{\}  

Supervisor Feedback: \{\}

\textbf{Consider the feedback above when planning your next action.}  

\textbf{ASSISTANT: TASK:}

\end{tcolorbox}

The following prompt shows how feedback is incorporated into the next reasoning–action iteration for MolmoACT~\cite{lee2025molmoact}:
\begin{tcolorbox}[breakable, colback=gray!5, colframe=gray!50, boxrule=0.5pt, left=2mm, right=2mm]

The task is \{TASK\}.  

What is the action that the robot should take?  

To figure out the action that the robot should take to \{TASK\}, please consider the following.

\textbf{Here is the previous reasoning and feedback:}  

- Previous Reasoning (Trace thought): \{\}  

- Supervisor Feedback: \{\}

Please take into account the supervisor's feedback above to improve your next reasoning and action decision.

Let's think through it step by step.  

First, what is the depth map for this image?  

Second, what is the trajectory of the end effector?  

Based on the depth map of the image and the trajectory of the end effector, what is the action that the robot should take?

\end{tcolorbox}

The following prompt shows how feedback is incorporated into the next reasoning–action iteration for ER-1~\cite{yuan2025embodied} for task ``Select grasp point'':
\begin{tcolorbox}[breakable, colback=gray!5, colframe=gray!50, boxrule=0.5pt, left=2mm, right=2mm]

Provide one or more points coordinate of objects region this sentence describes: \{TASK\}.  

The results are presented in a format \textless point\textgreater [[x1,y1], [x2,y2], ...] \textless /point\textgreater.  

\textbf{Previous Attempt:}  

Robot's Reasoning: \{\}  

Supervisor Feedback: \{\}

You FIRST think about the reasoning process as an internal monologue and then provide the final answer.  

The reasoning process and answer are enclosed within \textless think\textgreater  \textless /think\textgreater  and \textless answer\textgreater  \textless /answer\textgreater  tags.  

The answer consists only of several coordinate points, with the overall format being:  

\textless think\textgreater reasoning process here \textless /think\textgreater \textless answer\textgreater \textless point\textgreater [[x1, y1], [x2, y2], ...] \textless /point\textgreater \textless /answer\textgreater

\end{tcolorbox}

\paragraph{Prompts for Feedback Generation}
The following prompt is the prompt to generate the feedback:

\begin{tcolorbox}[breakable, colback=gray!5, colframe=gray!50, boxrule=0.5pt, left=2mm, right=2mm]
You are an expert robot task supervisor providing constructive feedback.

\textbf{Your Task:}

Analyze the current situation in the image and the robot's reasoning and action. 
Provide specific, actionable feedback:

1. What might be problematic with the current approach?

2. What should the robot focus on or prioritize?

3. What specific aspect needs adjustment?

\{ '4. How can the action quality (reward) be improved?' if reward\_info else '' \}

Provide concise, actionable feedback in 2-3 sentences. Be constructive and specific.

\textbf{Overall Objective:}

\{instruction\}

\textbf{Robot's Current Reasoning:}

\{reasoning\_text\}

\textbf{Evaluation Score:} \{score:.2f\} 

(Low score indicates potential issues with the current plan)

\{reward\_section\}

\end{tcolorbox}

\subsection{Reasoning Verifier}
\label{sec:re}
In our implementation, we adopt Qwen2.5-VL-7B~\cite{bai2025qwen2} as the reasoning verifier due to its strong multimodal perception and reasoning capabilities.
To accommodate diverse reasoning modalities in embodied decision-making, we categorize reasoning into three types:

\noindent\textbf{(1) Multimodal Reasoning.} 
As a representative of multimodal reasoning, E-CoT~\cite{zawalski2024robotic} integrates visual perception with textual task understanding to produce reasoning traces that effectively bridge high-level task goals and low-level actions. This type of reasoning aligns the spatial and functional semantics described in the task instruction with perceptual evidence from the current observation $o_t$, generating contextually grounded intermediate reasoning steps. In our implementation, the textual reasoning trace remains text, while visual perception results are rendered onto the image to construct a structured visual prompt. This design allows the reasoning verifier $V_{\text{c}}$ to assess cross-modal consistency and the plausibility of candidate plans. Each reasoning trace in E-CoT is organized hierarchically to capture the full decision-making process, including the overall \textit{TASK}, a decomposed \textit{PLAN}, individual \textit{SUBTASKS} with their corresponding \textit{SUBTASK REASONING}, and low-level action specifications such as \textit{MOVE}, \textit{MOVE REASONING}, \textit{GRIPPER POS}, and \textit{VISIBLE OBJECTS}. By representing both perceptual grounding and action-relevant reasoning in a structured format, E-CoT enables $V_{\text{c}}$ to evaluate reasoning traces for coherence, contextual alignment, and task feasibility, forming the backbone of our history-aware, feedback-guided verification framework.

\noindent\textbf{(2) Textual Reasoning.}  
Textual reasoning focuses on purely linguistic forms of reasoning, such as goal decomposition and temporal plan sequencing derived directly from task instructions. In this reasoning paradigm, high-level plans are represented in natural language, capturing the sequence of sub-goals and their dependencies without relying on perceptual grounding. The reasoning verifier $V_{\text{c}}$ assesses the logical coherence and task relevance of these textual plans through contextual embeddings, ensuring that each proposed step aligns with the overall task objective. Representative works such as ER-1~\cite{yuan2025embodied} produce \texttt{<think>} traces like \texttt{<think>I should first grasp the handle and lift it from the base...</think>}, while thinkACT generates stepwise instructions in natural language, e.g., \emph{``To put the strawberry in the drawer, the robot needs to: 1. pick the strawberry ...''}. By encoding purely linguistic reasoning in this structured manner, $V_{\text{c}}$ can evaluate the feasibility and consistency of high-level plans, forming a critical component in our history-aware, feedback-guided verification framework for sequential embodied tasks.

The prompt for evaluation of these two categories is shown as:
\begin{tcolorbox}[breakable, colback=gray!5, colframe=gray!50, boxrule=0.5pt, left=2mm, right=2mm]

You are an expert robot task supervisor.  

Your role is to determine if a robot's thinking process is sound for the current situation.

\textbf{Your Task:}  

Based on the \textbf{Overall Objective}, the \textbf{visual progression from historical to current observations}, the \textbf{Recent Action History} (if provided), and the \textbf{Current Robot's Internal Reasoning}, evaluate: Is the robot's current plan logical, coherent, and does it represent a promising step toward achieving the objective? Consider whether it builds upon or contradicts previous actions.

Answer with only the single word "yes" or "no".

\textbf{Overall Objective:}  

"\{instruction\}"

\{image\_description\}  

\{history\_context\}

\textbf{Current Robot's Internal Reasoning for its Next Action:}  

---  

\{ecot\_reasoning\_text\}  

---

\end{tcolorbox}

\noindent\textbf{(3) Spatial Reasoning.}  
Spatial reasoning focuses on trajectory-level and keypoint-based reasoning, where reasoning traces are converted into visual prompts that capture the intended spatial configuration of the robot within the scene. To guide the reasoning verifier $V_{\text{c}}$ in evaluating spatial plausibility, we construct representative examples exhibiting different degrees of alignment between the predicted intent and the scene geometry. High-confidence examples correspond to reasoning traces whose predicted spatial configurations closely match the observable environment, while low-confidence instances reflect misaligned or physically infeasible spatial grounding. This design enables $V_{\text{c}}$ to effectively discriminate between well-grounded and inconsistent spatial reasoning, ensuring robust assessment of spatial intent within the history-aware, feedback-guided framework. Representative works such as MolmoACT~\cite{lee2025molmoact} encode spatial reasoning with a combination of \emph{Depth Perception Tokens} (e.g., \texttt{<depth\_95>}) and \emph{Visual Reasoning Traces} specifying trajectory keypoints (e.g., \texttt{[[123,242],[245,163]]}), providing a structured representation that directly informs verifier scoring.

\begin{figure*}
\centering
\includegraphics[width=\linewidth]{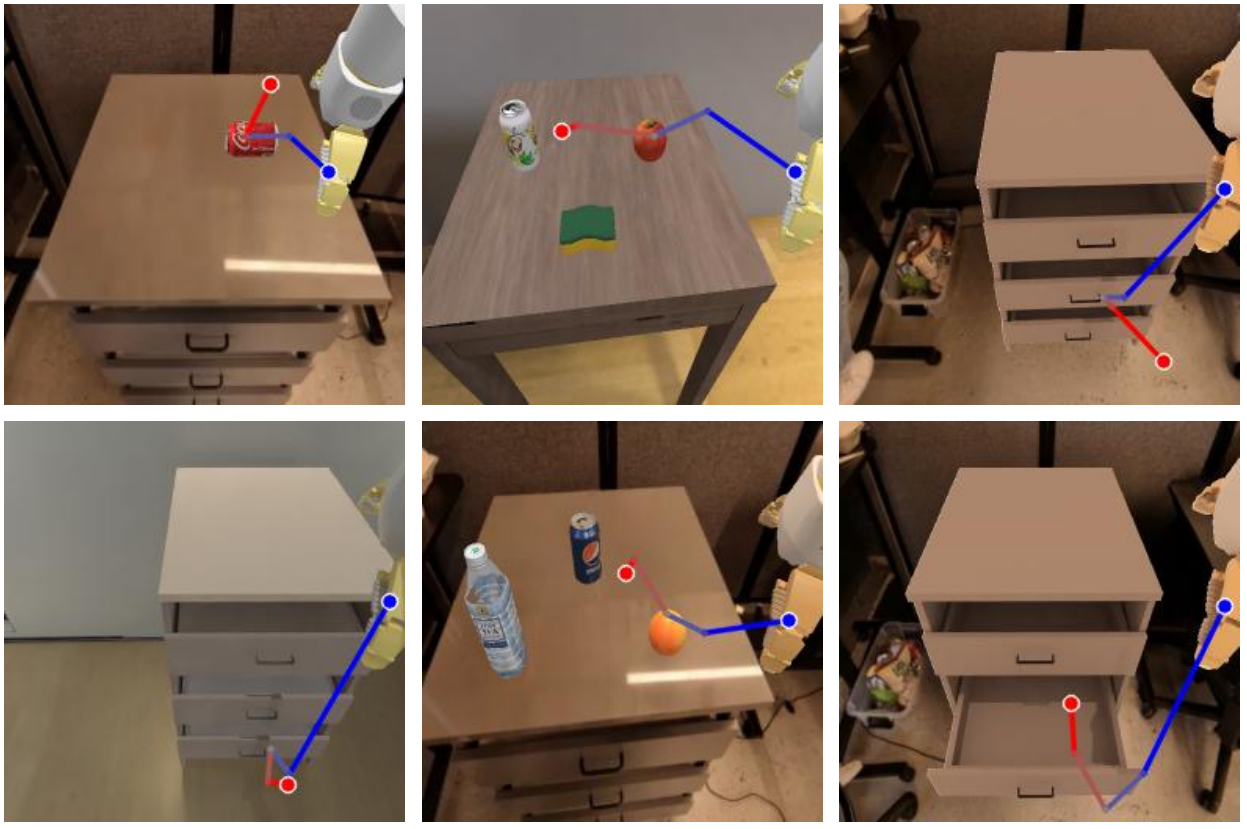}
\captionof{figure}{Visualization of spatial reasoning traces. 
  Each example shows the overlaid keypoints representing the predicted robot trajectory, 
  where blue and red dots indicate the start and end positions, respectively. 
  Well-aligned traces correspond to physically plausible reasoning, 
  while misaligned traces indicate inconsistent or infeasible spatial grounding.
  }
\label{fig:show_trace}
\end{figure*}

Concretely, we overlay the keypoints on the observed images, where the blue and red endpoints indicate the starting and ending positions of the trajectory, respectively, as illustrated in \cref{fig:show_trace}. 

The prompt for evaluation of spatial reasoning is:

\begin{tcolorbox}[breakable, colback=gray!5, colframe=gray!50, boxrule=0.5pt, left=2mm, right=2mm]

You are an expert robot task supervisor. Your role is to determine whether a robot's reasoning is correct for the given task.

\textbf{Your Task:}  

We draw the robot's reasoning on the image with points and lines. The trajectory visualization uses a blue-to-red color gradient to represent the temporal sequence: blue lines and points indicate the starting position and early stages of the robot's planning, while red lines and points represent the ending position and later stages. The gradient shows the chronological progression of the robot's end-effector path through time. Based on the \textbf{Overall Objective} and the current image, examine the \textbf{Robot's Reasoning}. Is the reasoning logical, coherent, and a promising step toward achieving the objective? Answer only "yes" or "no".

\textbf{Below are several examples for reference:}

\textbf{Example 1:}  

Instruction: \{instruction 1\}

Observation: \{image example 1\}  

Answer: yes

\textbf{Example 2:}  

Instruction: \{instruction 2\}

Observation: \{image example 2\}  

Answer: yes

\textbf{Example 3:}  

Instruction: \{instruction 3\} 

Observation: \{image example 3\}  

Answer: no

\textbf{Overall Objective:}  

"\{instruction\}"

\textbf{Current Observation:}  

\{image observation\}

\textbf{Recent History (optional):}  

For each historical observation and reasoning pair:  

Observation: \{historical image example\}  

Reasoning Trace: \{historical reasoning trace\}

\end{tcolorbox}

\subsection{Action Verifier}
\label{sec:act}
To evaluate candidate actions for each reasoning trace, we introduce a learned action-verifier architecture inspired by~\cite{kwok2025robomonkey}.

The verifier takes as input the current observation $o_t$, the task instruction $I$, and a sampled action $a$, and outputs a confidence score $S_a$ indicating the likelihood that the action will succeed under the given instruction.

The verifier is trained on a synthetic preference-based dataset of action comparisons derived from expert demonstrations. For each observation-instruction pair, multiple candidate actions are sampled from the policy, and a ground-truth error metric relative to the expert action is computed. Higher-quality actions are labeled as preferable. 
Following~\cite{kwok2025robomonkey}, the action verifier adopts LLaVA-7B~\cite{liu2024improved} as the backbone, replacing its final unembedding layer with a lightweight reward head that outputs scalar confidence scores. The action verifier is pretrained only on the same dataset as the base VLA policy, strictly excluding all evaluation environments and test tasks. 

For inference, for each candidate action $a$ under observation $o$, reasoning $c$ and instruction $I$, the verifier outputs
\begin{equation}
S_a = V_{\text{a}}(o, I, a)
\end{equation}
which is normalized to $\hat{S}_a$ and subsequently combined with the reasoning trace score to produce the joint score.

\subsection{Adaptive Online joint Selection}
\label{sec:select}
With dual verifiers for joint sample, each $J_t^{i,j} = (c_{t}^i, a_{t}^{i,j})$ is assigned with a score $S_{t}^{i,j}$ in a joint sample batch.
Although joint scoring provides a robust measure of reasoning–action consistency, naively evaluating all batches can be computationally expensive and may lead to redundant exploration, especially in long-horizon embodied tasks where the action space is large. To address this, we propose an \emph{adaptive online selection} strategy that dynamically balances exploration and exploitation during inference, which can be expressed as:

\begin{equation}
J_t^{\mathcal{B}_i} =
\begin{cases}
\operatorname{Random}(\mathcal{B}_t^i), & u \le \epsilon \\
J_t^{i,j^*}, & u > \epsilon \ \text{and}\  S_t^{i,j^*} > \eta \\
\text{next batch } \mathcal{B}_t^{i+1}, & u > \epsilon \ \text{and}\  S_t^{i,j^*} \le \eta
\end{cases}
\end{equation}
where $J_t^{\mathcal{B}_i}$ is the selected reasoning–action pair from batch $\mathcal{B}_t^i$ at timestep $t$, $u \sim \text{Uniform}(0,1)$ determines exploration vs.\ exploitation, $\epsilon$ is the exploration probability, $J_t^{i,j^*}$ is the highest-scoring pair with joint score $S_t^{i,j^*}$, $\eta$ is the quality threshold, and $\mathcal{B}_t^{i+1}$ is the next candidate batch.

This design reflects several principled considerations. First, the multiplicative joint score explicitly enforces that both reasoning and action quality are high, which is crucial in embodied tasks where a good plan alone is insufficient if the corresponding action is infeasible. Second, the $\epsilon$-greedy policy encourages the system to explore alternative reasoning–action hypotheses, preventing premature convergence to suboptimal trajectories. Finally, the threshold $\eta$ filters out low-confidence candidates while dynamically allocating computational resources to the most promising pairs. Together, these design choices enable the framework to achieve efficient, robust, and adaptive selection of reasoning–action pairs during test-time scaling, minimizing redundant evaluation while maintaining high task performance.

\begin{figure}[htbp]
    \centering
    \includegraphics[width=0.95\textwidth]{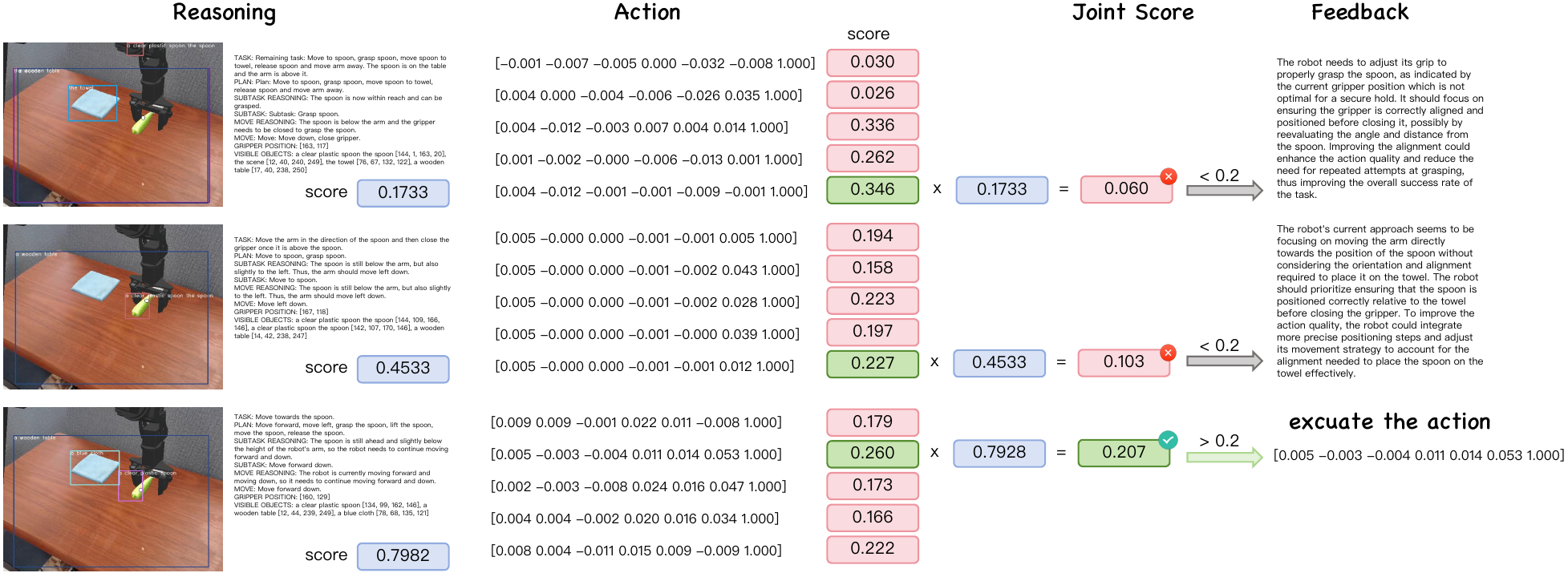} 
    \caption{Example of the selection process. The figure demonstrates three iterations. The first two are rejected due to incomplete object detection (missing the spoon and the cloth, respectively), while the third successfully identifies all objects and achieves the highest joint score for execution.}
    \label{fig:reasoning_process}
\end{figure}

The qualitative results of the joint score and selection process are visualized in Fig.~\ref{fig:reasoning_process}. As shown in Fig.~\ref{fig:reasoning_process}, the model generates multiple joint sample pairs, each associated with a joint score. In the first reasoning instance, the model exhibits a significant perception failure by failing to recognize the clear plastic spoon within the scene. Consequently, the generated move reasoning lacks a concrete target, leading to a low joint score and the rejection of the action. In the second iteration, although the model successfully detects the spoon, it overlooks the blue cloth (the target placement area). In contrast, the third reasoning attempt achieves the highest joint score (0.7982) by correctly perceiving all critical environmental entities, including the spoon and the blue cloth with their respective spatial coordinates. The successful execution of this selected action demonstrates that robust object detection is a prerequisite for logically consistent robotic manipulation.

\subsection{Examples on Feedback-Guided Iterative Refinement}
\label{sec:example_feedback}
\input{table/appendix_feedback_case}
We illustrate the process of feedback-guided iterative refinement using a concrete task instance where the robot must move a spoon onto a towel in~\cref{tab:fgir_feedback}. 
The procedure alternates between single-sample generation and verified joint reasoning-action sampling. Initially, the system generates candidate reasoning traces and actions without verification in Steps 0-9. 

Starting from Step 10, we introduce the pruning and verification mechanism. For each batch, a single reasoning trace is sampled and multiple candidate actions are generated. Each action is scored using the joint reasoning-action score, which combines the trace plausibility with the predicted action reward. Samples falling below the acceptance threshold are pruned, and targeted feedback is generated to guide the next iteration. For example, early iterations reveal that the robot incorrectly identifies the target spoon among multiple visible objects, resulting in low joint scores (0.0769-0.0515) and triggering corrective feedback. By the fourth try at step 10, the system begins accepting samples with higher joint scores (0.2002) and start to generate reasoning and action for next timestep. 

However, we also observe that the score drops when conducting the second sampling, primarily because the feedback perturbs the model’s immediate policy and pushes it to explore alternative reasoning–action trajectories that are not yet optimized. Despite this short-term degradation, such feedback-driven exploration is beneficial, as it steers the model toward more accurate object understanding and ultimately leads to higher-quality samples in later iterations.

This example demonstrates the effectiveness of iterative refinement with feedback: low-quality actions are pruned, corrective guidance is provided, and reasoning-action consistency gradually improves, enabling robust task execution without exhaustive search.

\subsection{Example of Feedback-Guided Iterative Refinement with Supervisor Feedback}
\label{sec:example_Supervisor}
We illustrate a key iteration in our Feedback-Guided Iterative Refinement process for the task: \textit{put the spoon onto the towel}. The robot generates a reasoning sample, evaluates multiple candidate actions, and prunes samples below the joint score threshold. Supervisor feedback is then generated to guide subsequent batches. An example of timestep 10 is shown in \cref{tab:fgir_feedback}.

\paragraph{Feedback Example:}  
For instance, in step 10-2 in \cref{tab:fgir_feedback}, the supervisor feedback was:  
\begin{quote}
\textit{``The robot seems to be focusing on moving towards the spoon but hasn't yet grasped it correctly. It should prioritize ensuring that the gripper is properly aligned with the spoon before attempting to pick it up. A minor adjustment could be refining the grip position to make sure it aligns more directly with the spoon's placement. This should help improve the action quality by making the next move more precise and effective.''}
\end{quote}

This feedback is incorporated into the next reasoning generation, guiding the robot to adjust its reasoning and improve subsequent actions. By iteratively combining reasoning evaluation, action verification, and supervisor feedback, the robot progressively achieves more precise and effective manipulation.

\section{Evaluation Details in Simulation}

\subsection{Implementation}
\label{sec:impl}
All experiments use the same inference and verification pipeline with a small set of task-level hyperparameters selected to balance computation and sample diversity. The experiment is performed three times to obtain the average result. For the E-CoT~\cite{zawalski2024robotic} (E-TTS) configuration reported in the main paper we use: we evaluate every 10 steps , exploration probability $\epsilon=0.1$, and a threshold $\eta=0.4$. Reproducibility is ensured by fixing the random seed to $42$. To limit runtime, for each batch, we sample 20 actions, and the maximum batch number is 20. The verifier maintains a short history window of size $k=3$ (most recent reasoning–action pairs) to provide temporal context during scoring.

We evaluate two comparative base models with small, controlled modifications to the above hyperparameters. For MolmoAct~\cite{lee2025molmoact} we retain the E-CoT configuration except that the maximum number of batches is reduced to $m=25$. For ER-1~\cite{yuan2025embodied} we set a more conservative sampling budget and threshold: $m=20$ and $\eta=0.3$. These variant settings are chosen to respect each baseline's computational profile while keeping the overall experimental comparison fair.

In our experiments, the naive test-time scaling (TTS) baseline employs a \emph{best-of-N} strategy~\cite{lightman2023let}. Specifically, for each sampling timestep, the model generates $N$ candidate solutions by independently scaling both the reasoning and action components. Each candidate is then evaluated using the same vision-language foundation model as in our approach, and the one with the highest verification score is selected as the final output. This strategy aims to improve performance by leveraging multiple attempts, without incorporating the more sophisticated reasoning or planning mechanisms present in our method.

Formally, for timestep $t$, given $N$ candidate reasoning-action pair $\{ J_t^1, \dots, J_t^N \}$ and their corresponding verification scores $\{ S_1, S_2, \dots, S_N \}$, the selected pair $J_t^*$ is:
\[
J_t^* = \arg\max_{i \in \{1, \dots, N\}} S_i.
\]

This simple approach serves as a strong naive baseline for comparison. The $N$ is set to be the average number of our sampling iterations, for E-CoT $N = 15$, for MolmoAct $N=5$, and for ER-1 $N = 5$.

For RoboMonkey, at each timestep $t$, it first generates a set of $\hat{N}$ candidate actions $\hat{A}$ from VLA models and fit a multivariate Gaussian distribution $\mathcal{N}(\mu_t, \Sigma_t)$. A refined set of $\hat{K}$ new actions $\tilde{A}$ is then resampled from this fitted proposal distribution, each refined action is evaluated by a reward model $R$, and the action that maximizing the reward is selected for execution. Here $N$ and $K$ for E-CoT is 2 and 10.

\subsection{Base Models}
\label{sec:base_model}
\paragraph{\textbf{E-CoT}} 
In our experiments we adopt a base model E-CoT~\cite{zawalski2024robotic}, in which a pretrained vision-language backbone is fine-tuned to map image observations and language instructions directly to robot actions. We use this architecture as the “base model” for comparative evaluation, without the extra training or finetuning. Specifically, the base model takes as input the instruction string \(I\) and the current image observation \(o_t\), generates the reasoning $c_t$ and predicts an action \(a_t\) in an autoregressive paradigm.
\noindent The reasoning trace of E-CoT can be categorized as multimodal reasoning, which contains the following core types:  
\begin{itemize}
  \item \textbf{TASK:} a restatement or refinement of the original instruction.  
  \item \textbf{PLAN:} a high-level sequence of sub-tasks designed to achieve the goal.  
  \item \textbf{SUBTASK:} the specific sub-task selected for the current decision step.  
  \item \textbf{SUBTASK REASONING:} the justification or rationale for choosing the current sub-task given the scene and instruction.  
  \item \textbf{MOVE:} the primitive motion command (e.g., “move forward”, “turn left”) to advance toward the sub-task goal.  
  \item \textbf{MOVE REASONING:} the explanation of why this particular motion was selected, grounded in observations or robot state.  
  \item \textbf{GRIPPER POS:} the predicted or referenced end-effector position (in image or world coordinates) at the time of grasping or manipulation.  
  \item \textbf{VISIBLE OBJECTS:} a list of named objects and their bounding boxes or spatial descriptions detected in the scene, supporting grounded reasoning.  
\end{itemize}

These reasoning categories enable the policy to first reason about the task and then to understand the scene via object bounding boxes and gripper positions before predicting a low-level action, from goal to plan to detailed motion.

\paragraph{\textbf{ER-1}} 
We include Embodied-R1 (ER-1)~\cite{yuan2025embodied} as one of our base models. 
Embodied-R1 is a 3B Vision-Language Model trained to integrate spatial reasoning and sequential manipulation through a textual reasoning. 
ER-1 uses ``pointing'' as a unified, embodiment-agnostic intermediate representation. This point-centric approach serves as a bridge, translating the model's high-level perceptual understanding into a compact spatial signal that guides low-level action execution.

Architecturally, Embodied-R1 is built upon the Qwen2.5-VL model~\cite{bai2025qwen2} and is fine-tuned to master four distinct pointing abilities: \textit{Referring Expression Grounding (REG)} for object localization, \textit{Region Referring Grounding (RRG)} for specifying placement locations, \textit{Object Functional Grounding (OFG)} for identifying affordances, and \textit{Visual Trace Generation (VTG)} for planning object-centric trajectories. A key design choice is the training methodology. Instead of conventional Supervised Fine-Tuning (SFT), the authors employ a two-stage Reinforced Fine-tuning (RFT) curriculum. This paradigm, powered by the custom-curated \texttt{Embodied-Points-200K} dataset and multi-task reward functions, is specifically designed to overcome the ``multi-solution dilemma'' inherent in pointing tasks, where multiple coordinates can be valid answers. The generated points are ultimately consumed by a downstream action executor, enabling modular and generalizable zero-shot control.

Given a natural-language instruction and scene observation, the model first generates an internal reasoning trace in the format:
\begin{quote}
\texttt{<think>} I should first grasp the handle and lift it from the base. 
The drawer is on the tabletop, positioned further back, closer to the wall. 
I need to avoid obstacles and carefully move the moka pot from the current position 
to the right side of the drawer. \texttt{</think>}
\end{quote}
This text reasoning explicitly captures spatial relations and task sequencing, forming an interpretable intermediate step between perception and control. 
The final action prediction is then conditioned on both the visual input and the generated reasoning trace, aligning low-level motor control with high-level task intent.
 
ER-1 first generates the text reasoning then predicts both object affordance points and target region points using the CuRobo motion planner~\cite{sundaralingam2023curobo}, followed by zero-shot deployment on the WidowX robotic arm for Simpler evaluation. 

\paragraph{\textbf{MolmoAct}} 
While many existing VLA models typically map perception and instructions directly to action, MolmoAct~\cite{lee2025molmoact} introduces a novel approach by integrating perception, planning, and control through a structured reasoning pipeline, explicitly in space. This involves generating sequences of depth perception tokens, visual reasoning traces, and finally, precise action tokens, ensuring spatially grounded and explainable behaviors.
Specifically, the model employs depth perception tokens to enable 3D understanding, which is crucial for tasks requiring spatial awareness. Furthermore, MolmoAct generates 2D visual reasoning traces, representing planned end-effector trajectories, which directly align visual inputs with control outputs. This approach contrasts with methods that distill complex 3D trajectories into linguistic descriptions, which can lead to information loss. By conditioning each reasoning stage on the preceding outputs, MolmoAct ensures that the final actions are robustly grounded in both inferred depth and planned motion.

Given a natural-language instruction and scene observation, the model first generates an internal reasoning trace in the format:
\begin{quote}
The depth map of the image is:

\texttt{<DEPTH\_START> <DEPTH\_26> <DEPTH\_56> <DEPTH\_75> <DEPTH\_101> \\
<DEPTH\_69> <DEPTH\_55> <DEPTH\_25> <DEPTH\_69> <DEPTH\_8> \\ 
<DEPTH\_32> <DEPTH\_101> <DEPTH\_125> <DEPTH\_26> <DEPTH\_75> \\
<DEPTH\_25> <DEPTH\_55> <DEPTH\_25> <DEPTH\_9> <DEPTH\_69> \\
<DEPTH\_32> <DEPTH\_26> <DEPTH\_74> <DEPTH\_116> <DEPTH\_84>\\
<DEPTH\_75> <DEPTH\_74> <DEPTH\_116> <DEPTH\_116> <DEPTH\_9>\\
<DEPTH\_69> <DEPTH\_26> <DEPTH\_116> <DEPTH\_74> <DEPTH\_9>\\
<DEPTH\_125> <DEPTH\_74> <DEPTH\_25> <DEPTH\_116> <DEPTH\_9>\\
<DEPTH\_69> <DEPTH\_26> <DEPTH\_84> <DEPTH\_28> <DEPTH\_25> \\
<DEPTH\_8> <DEPTH\_25> <DEPTH\_116> <DEPTH\_25> <DEPTH\_9>\\
<DEPTH\_8> <DEPTH\_101> <DEPTH\_55> <DEPTH\_103> <DEPTH\_25>\\
<DEPTH\_74> <DEPTH\_56> <DEPTH\_9> <DEPTH\_116> <DEPTH\_9>\\
<DEPTH\_32> <DEPTH\_75> <DEPTH\_116> <DEPTH\_9> <DEPTH\_74> \\
<DEPTH\_28> <DEPTH\_32> <DEPTH\_8> <DEPTH\_69> <DEPTH\_69>\\
<DEPTH\_69> <DEPTH\_75> <DEPTH\_26> <DEPTH\_32> <DEPTH\_55>\\
<DEPTH\_74> <DEPTH\_8> <DEPTH\_56> <DEPTH\_8> <DEPTH\_8> \\
<DEPTH\_8> <DEPTH\_75> <DEPTH\_26> <DEPTH\_8> <DEPTH\_74> \\
<DEPTH\_74> <DEPTH\_9> <DEPTH\_56> <DEPTH\_101> <DEPTH\_56>\\
<DEPTH\_56> <DEPTH\_69> <DEPTH\_8> <DEPTH\_84> <DEPTH\_76>\\
<DEPTH\_103> <DEPTH\_30> <DEPTH\_69> <DEPTH\_55> <DEPTH\_25> \\
<DEPTH\_116> <DEPTH\_END>.}

The trajectory of the end effector in the image is:

\texttt{[[108,59], [81,89], [76,111], [76,125], [78,71]].}
\end{quote}

\paragraph{\textbf{$\pi_{0.5}$}}
$\pi_{0.5}$ is a hierarchical vision-language-action (VLA) model built on $\pi_0$, which achieves open-world generalization for robotic manipulation. Initialized from a web-trained VLM, the model first undergoes pre-training on a mixed dataset with FAST tokenizer for discrete action representation, then is post-trained for mobile manipulation with flow matching-based action expert to generate continuous actions, incorporating human instruction data to optimize high-level subtask inference. The model unifies high-level semantic reasoning and low-level action prediction in a single transformer architecture, fusing discrete and continuous action representations with a combined loss function, and leverages multi-source knowledge transfer to enable generalization to unseen home environments. At inference time, $\pi_{0.5}$ first infers a high-level semantic subtask from the global language prompt, robotic visual observations and proprioceptive states via its unified transformer backbone, then conditions the dedicated action expert on this predicted subtask to generate fine-grained continuous low-level action chunks for end-to-end mobile manipulator control.

Given a natural-language instruction and scene observation, the model first generates an internal reasoning trace in the format:
\begin{quote}
Pick up the pillow.
\end{quote}

\subsection{Details on Evaluation on SimplerEnv}
\label{sec:evaluate_simpler}
We conduct our experiments within the \texttt{SimplerEnv}~\cite{li24simpler} simulation suite, an open-source real-to-sim benchmark designed to evaluate generalist manipulation policies under standardized and reproducible conditions~\cite{li24simpler}. 
SimplerEnv provides two complementary evaluation regimes, \emph{visual matching}, which aligns simulated scenes with real-world appearance, and \emph{variant aggregation}, which systematically varies scene configurations (lighting, distractors, and textures) to assess robustness. 
Using SimplerEnv allows rapid and consistent benchmarking of embodied reasoning-action models, while maintaining high correlation with real-world robotic behavior.  
For the three representative work, we test our work for E-CoT~\cite{zawalski2024robotic} and Embodied-R1~\cite{yuan2025embodied} on SimplerEnv simulation on the widowx robot setup. For Molmoact~\cite{lee2025molmoact}, we test E-TTS on SimplerEnv simulation on th google robot setup.
To ensure fair and reproducible comparison with other method, our evaluation follows the official SimplerEnv setup for Embodied-R1 and Molmoact. As for E-CoT, given that the performance on SimplerEnv has not be reported in the paper, we only set the max timestep as 500 to fully evaluate its ability, with others following the default setting.
We report the \textbf{success rate} as the primary evaluation metric, computed as the ratio of successful trials to total trials described below for each task, and averaged across all tasks under both visual-matching and variant-aggregation settings. 

\paragraph{Google Robot.}
We follow the standard evaluation suite consisting of four representative language-conditioned manipulation tasks:
(1) \emph{pick coke can}, where the robot grasps and lifts a coke can placed at diverse tabletop positions and orientations, yielding 75 trials in total.
(2) \emph{move \{obj1\} near \{obj2\}}, involving 8 household objects and multiple spatial layouts to test relational reasoning, yielding 60 trials in total.
(3) \emph{open/close (top, middle, bottom) drawer}, which assesses articulated-object control from varying robot poses, yielding 54 trials in total. and 
(4) \emph{open top drawer; place apple into top drawer}, a long-horizon compositional task that combines sequential subgoals, yielding 27 trials in total.. 
Each configuration follows the official grid-based sampling of initial robot and object poses.

\paragraph{WidowX Robot.}
For the WidowX robot, we evaluate four tabletop manipulation tasks:
(1) \emph{put the spoon on the towel}, 
(2) \emph{put carrot on plate}, 
(3) \emph{put eggplant into yellow basket}. 
Each task is evaluated by 24 trials.

\subsection{Details on Evaluation on LIBERO}
\label{sec:evaluate_libero}
We evaluate MolmoAct~\cite{lee2025molmoact} on the LIBERO simulation benchmark on Franka Emika Panda arm, a comprehensive multi-task benchmark designed to assess the capabilities of robotic manipulation systems.
Each demonstration includes front-view and wrist-view RGB images, natural language task instructions, and delta end-effector pose trajectories.
We follow~\cite{lee2025molmoact} and evaluate E-TTS on four different task, \textbf{LIBERO-Spatial}, \textbf{LIBERO-Object}, \textbf{LIBERO-Goal}, and \textbf{LIBERO-Long}, each comprising 10 task categories with 500 expert demonstrations.
For fair comparison, we set the action chunk with a fixed window size of 8 for MolmoAct.

\subsection{Experiments on E-CoT}
\label{sec:ecot}
\begin{figure*}
\centering
\includegraphics[width=\linewidth]{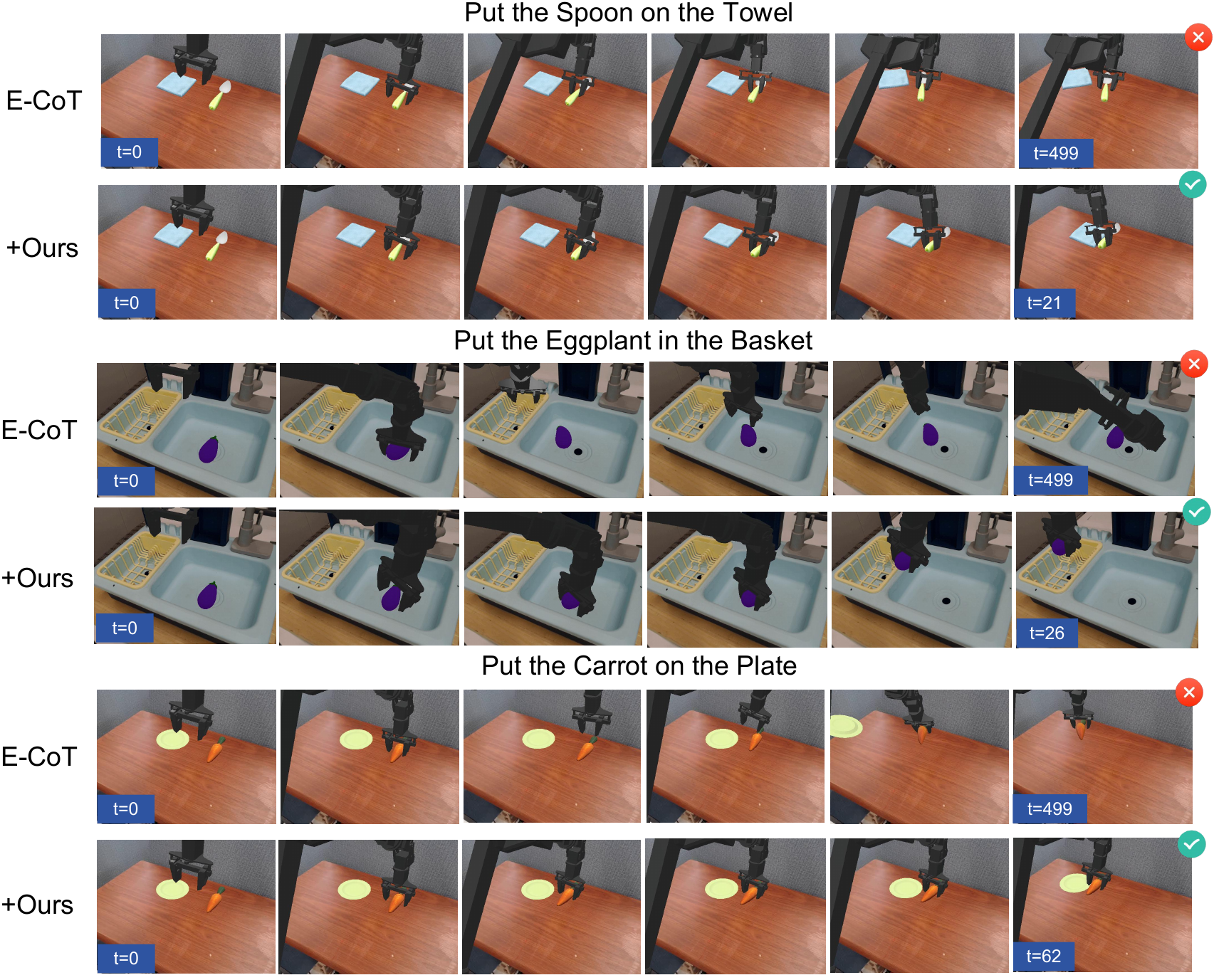}
\captionof{figure}{Qualitative comparison between E-CoT and E-CoT+Ours (E-TTS) on SimplerEnv across three manipulation tasks:``Put the Spoon on the Towel,”, ``Put the Eggplant in the Basket,” and ``Put the Carrot on the Plate.” Each row shows temporal execution frames. While E-CoT often fails to complete the placement action, our approach achieves precise object manipulation with shorter completion time and higher success rate.}
\label{fig:ecot_app}
\end{figure*}

\begin{figure*}
\centering
\includegraphics[width=\linewidth]{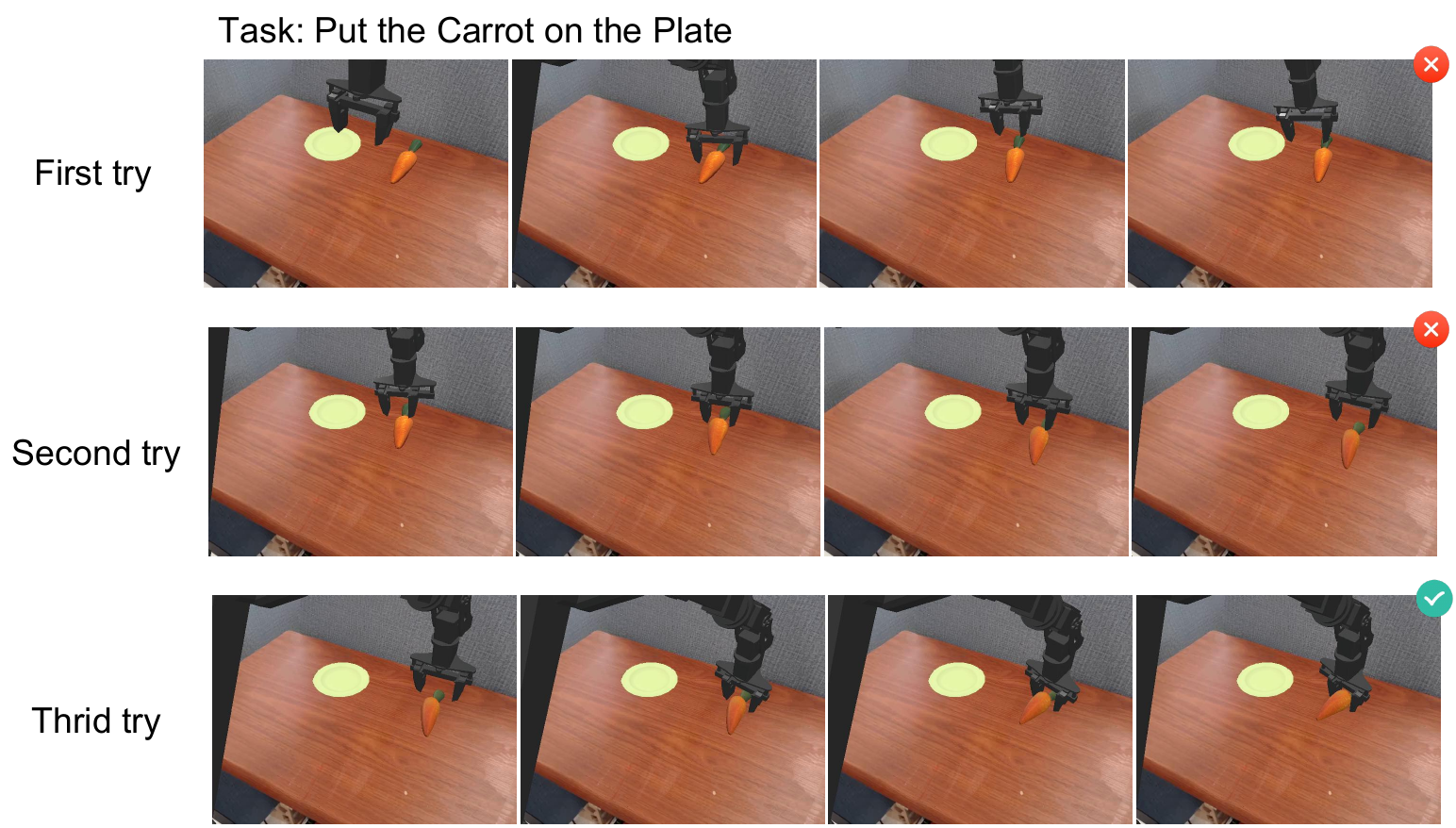}
\captionof{figure}{Visualization of self-corrective behaviors exhibited by \textbf{E-CoT + Ours} in the task 
  \textit{“Put the carrot on the plate.”} Each row shows consecutive attempts from the same rollout. 
  After failing to align the carrot with the plate in the first and second tries, 
  the model re-evaluates its reasoning and action candidates through closed-loop feedback 
  and successfully completes the task in the third attempt. 
  This demonstrates the effectiveness of our history-aware joint reasoning–action scaling 
  in enabling adaptive correction during sequential manipulation.}
\label{fig:ecot_app_correction}
\end{figure*}

\input{table/app_robo_ours}

We present qualitative comparisons between our method and the basemodel E-CoT~\cite{zawalski2024robotic} on three representative manipulation tasks. As shown in~\cref{fig:ecot_app}, E-CoT frequently exhibits inaccurate grasping or misalignment between the object and the target surface, resulting in task failure even after extended time steps. In contrast, with E-TTS it can successfully accomplishes all tasks with clear spatial alignment and stable placement behaviors. The performance improvement primarily stems from our enhanced reasoning-action coupling mechanism, which enables the model to better interpret spatial relations and dynamically adjust motion trajectories. Furthermore, with E-TTS, E-CoT achieves faster convergence (e.g., $t=21$ vs. $t=499$) and more consistent success across diverse object geometries and environments, demonstrating superior robustness and generalization capability in embodied manipulation.

As shown in Table~\ref{tab:performance_comparison_robo_ours}, our method outperforms the baseline Robomonkey across all scenarios. Specifically, we achieve a 51.5\% relative improvement in average success rate (from 26.38\% to 39.81\%) with only a marginal 7.8\% increase in average latency. Remarkably, in specific tasks such as put carrot on the plate (denoted as ``Carrot''), \methodName{} even reduces the average inference time (12.49s vs. 12.81s). This performance-efficiency trade-off highlights the effectiveness of \methodName{} without the typical burden of heavy computational overhead.

We further find that there is a self-correction capability of \textbf{E-CoT + Ours} in embodied manipulation. 
As illustrated in~\cref{fig:ecot_app_correction}, the robot initially fails to place the carrot on the plate due to suboptimal grasp or trajectory prediction. 
However, instead of terminating the episode, the model leverages historical observations stored in the buffer to refine its reasoning and resample improved actions through the verifier’s feedback. 
This iterative refinement enables the model to recover from previous failures and ultimately succeed in later attempts without any external supervision. 
Such closed-loop adaptability is absent in conventional test-time scaling methods, highlighting the benefit of jointly scaling reasoning and actions with history-aware feedback in embodied tasks.

\subsection{Experiments on Embodied-R1}
\label{sec:er1}
\begin{figure*}
\centering
\includegraphics[width=\linewidth]{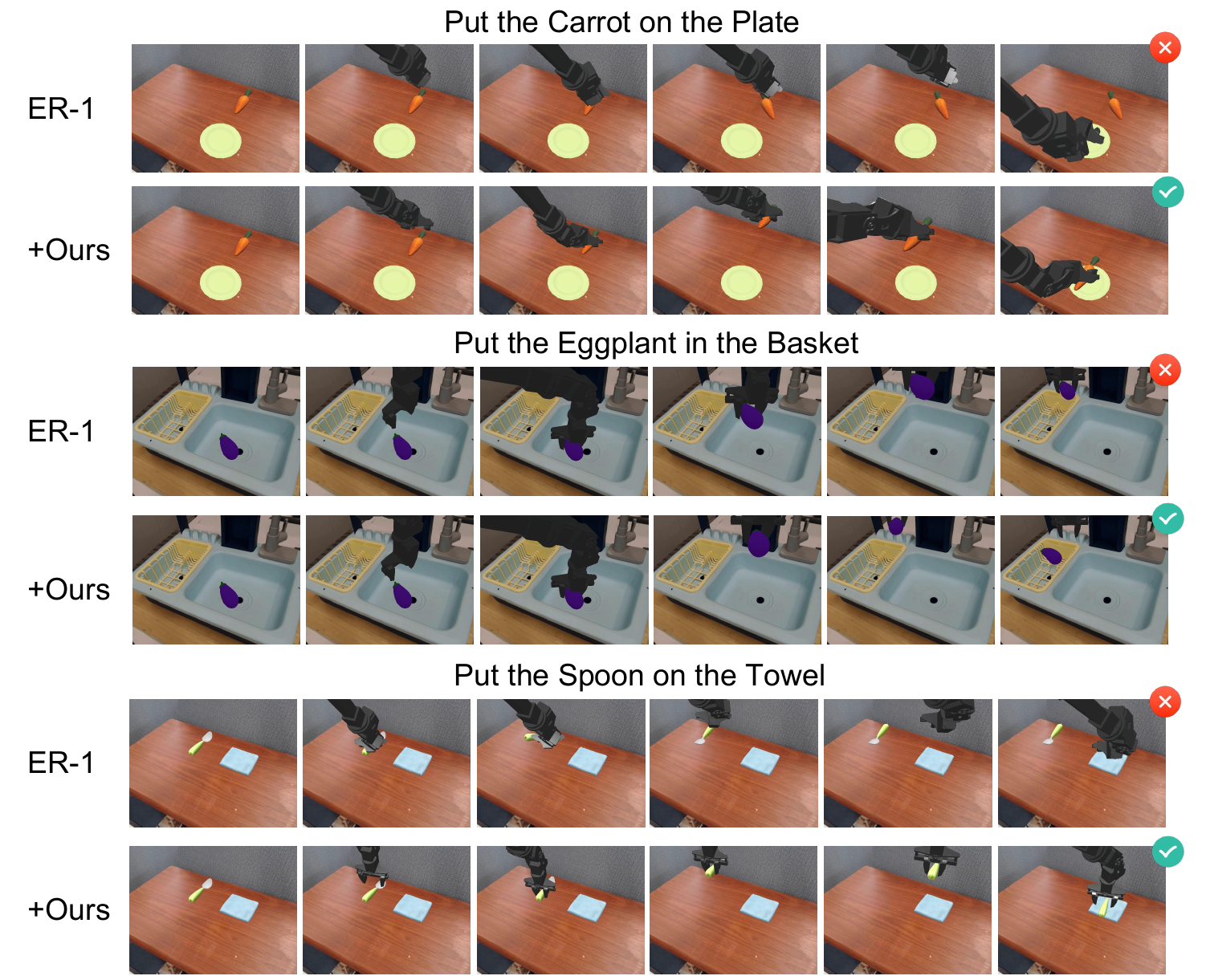}
\captionof{figure}{Qualitative comparison between ER-1 and ER-1+Ours (E-TTS) on SimplerEnv on three manipulation tasks: ``Put the Carrot on the Plate,” ``Put the Eggplant in the Basket,” and “Put the Spoon on the Towel.” While ER-1 often fails to complete fine-grained placement due to inaccurate spatial reasoning, our E-TTS-enhanced model achieves more consistent object alignment and successful task completion.}
\label{fig:er1_app}
\end{figure*}

\input{table/appendix_er1}

We evaluate the effectiveness of our proposed E-TTS module on the Simpler WidowX benchmark,
comparing it with the baseline ER-1~\cite{yuan2025embodied} and a naive test time scaling (TTS) variant.
As shown in~\cref{fig:er1_app} and~\ref{tab:simpler_result_append_er1}, ER-1 struggles with tasks involving precise spatial reasoning and stable object placement, often resulting in partial or failed completions.
The naive TTS approach provides limited improvement, indicating that merely augmenting textual reasoning is insufficient for robust execution.

In contrast, our E-TTS method significantly enhances manipulation success rates across all tasks, particularly in scenarios such as ``Put the Eggplant in the Basket,” where the success rate doubles compared to ER-1.
This improvement stems from E-TTS’s ability to dynamically refine action representations through reasoning-guided trajectory sampling and verification, allowing the model to interpret contextual cues and correct motion errors during execution.
Overall, E-TTS not only improves task completion accuracy by over 5\% on average but also produces smoother and more human-like manipulation behaviors, demonstrating its strong generalization and control capability in embodied environments.

\subsection{Experiments on MolmoAct}
\label{sec:molmoact}
\begin{figure*}
\centering
\scalebox{0.83}{
\includegraphics[width=\linewidth]{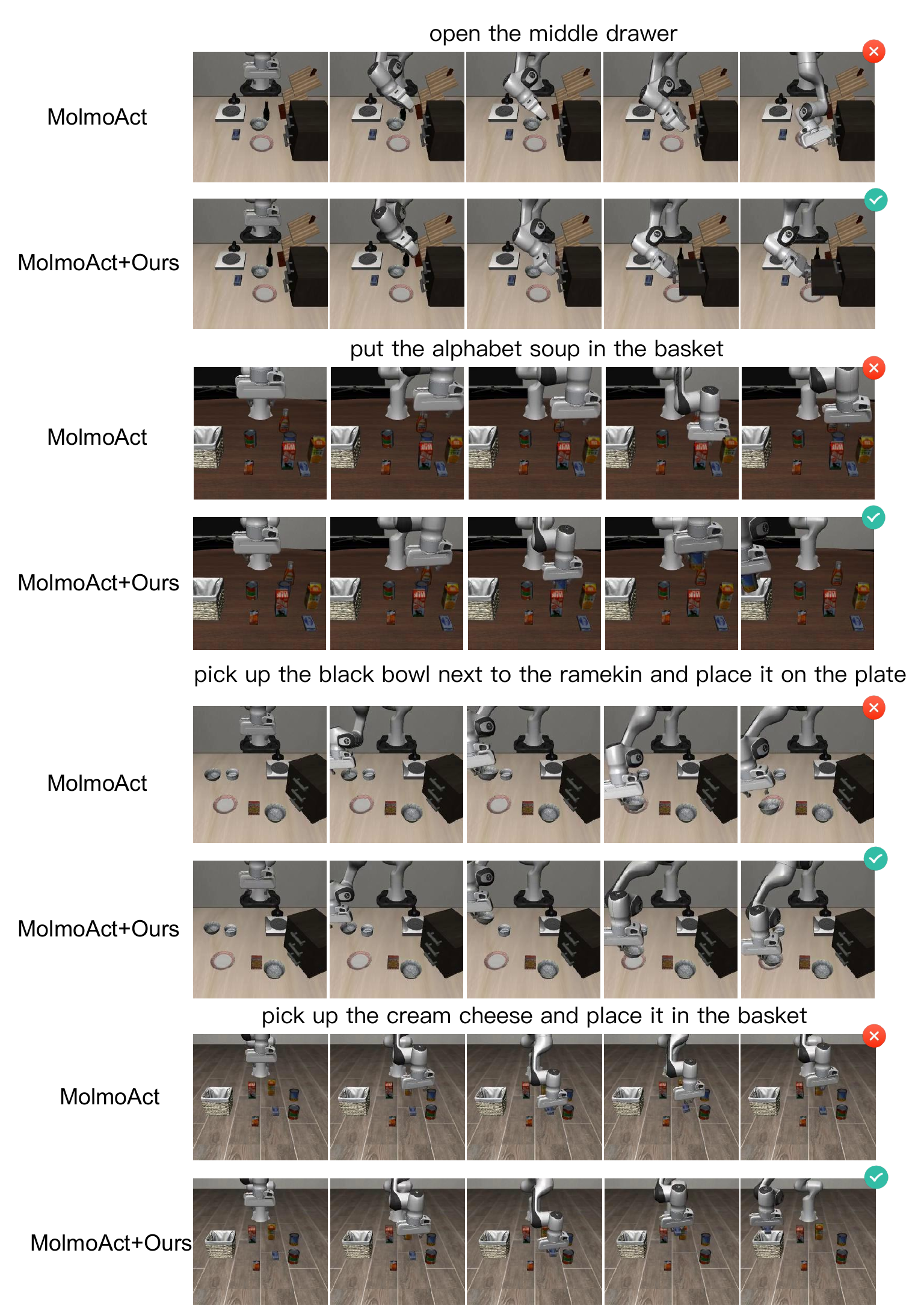}
}
\captionof{figure}{Qualitative comparison between MolmoAct and MolmoAct+Ours on LIBERO on four manipulation tasks: ``Open the Middle Drawer,” ``Put the Alphabet Soup in the Basket,” ``Pick up the Black Bowl Next to the Ramekin and Place it on the Plate,” and ``Pick up the Cream Cheese and Place it in the Basket.” While MolmoAct often fails to accomplish fine-grained actions or misinterprets spatial relations, our method consistently completes each task with accurate grasping, correct placement, and improved temporal efficiency.}
\label{fig:molmoact_app}
\end{figure*}

\begin{figure*}
\centering
\includegraphics[width=\linewidth]{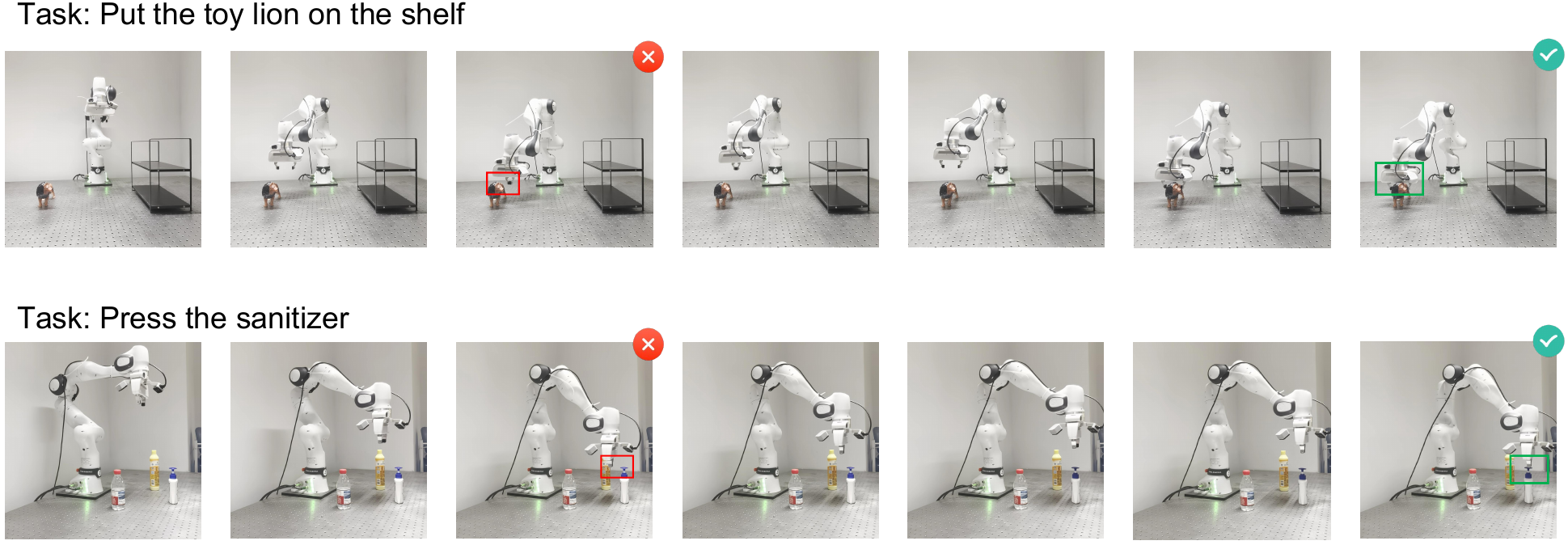}
\captionof{figure}{Visualization of real-world rollouts on two representative tasks: Put the toy lion on the shelf (top) and Press the sanitizer (bottom). Each row illustrates sequential observations from the robot during execution, where green boxes denote successful interactions and red boxes indicate failed attempts. The results highlight that basemodel + E-TTS can recover from previous failures and successfully complete the task through self-corrective behaviors.}
\label{fig:real_rollout_app}
\end{figure*}

\input{table/app_vlabench}

\input{table/app_real_data}

\input{table/notation}

We conduct qualitative comparisons to evaluate the effectiveness of our proposed improvements on top of the MolmoAct~\cite{lee2025molmoact} framework. As illustrated in~\cref{fig:molmoact_app}, MolmoAct often struggles with tasks requiring precise spatial reasoning, such as correctly grasping the target object or aligning it with the target region. For example, in the``Open the Middle Drawer” and ``Pick up the Cream Cheese” tasks, the baseline fails to execute the final step due to incomplete motion planning and inaccurate affordance understanding. In contrast, with E-TTS, it successfully completes all tasks with stable manipulation trajectories and consistent temporal progress. The improvement arises from our adaptive reasoning–action sampling and verification module, which dynamically refines spatial representations and optimizes low-level motor control based on multimodal feedback. This enhancement not only yields more natural and interpretable behaviors but also significantly reduces the number of required time steps, highlighting the superior efficiency and robustness of our approach in complex embodied manipulation scenarios.

\subsection{Experiments on $\pi_{0.5}$}
\label{appendix:pi05}

As illustrated in the complete results, integrating our method with the base $\pi_{0.5}$ model yields consistent performance gains across diverse task compositions. Specifically, in Track 1, our approach achieves a notable improvement in the average success rate from $0.39$ to $0.42$, with significant boosts in semantic-heavy tasks such as \textit{select\_poker} (SPo), where the success rate increases from $0.46$ to $0.63$.

We additionally compare E-TTS with TACO, a recent test-time scaling framework using pseudo-count estimation. As shown in Table~\ref{tab:taco}, E-TTS achieves a better average success rate, validating its performance. 

\input{table/taco}

\section{Real-world Experiments}
\label{appendix:real_world}
We collect a real-world dataset of 400 manipulation episodes. Each trajectory is recorded at 20\,Hz with synchronized third-person and wrist-mounted RGB streams (376$\times$672 and 480$\times$640 resolution, respectively) alongside 7-DoF end-effector actions, timestamps, and task annotations. The corpus spans four everyday manipulation goals, with 100 demonstrations per task and a total of 17{,}778 labeled frames, as shown in~\cref{tab:dataset}.

To evaluate the model's performance, we design four distinct manipulation tasks, which cover non-pick-and-place tasks, long-horizon tasks, and deformable object manipulation. ``Lion'' refers to ``put the toy lion on the shelf'', ``Bag'' refers to ``hang the bag on the shelf'', ``Sanitizer'' refers to ``press the sanitizer'' and ``Drawer'' refers to ``put the coke can on the upper drawer''. 

Each task captures distinct contact dynamics and workspace layouts, ensuring coverage over both short-horizon button-press behaviors and long-horizon pick-and-place maneuvers. This diversity, combined with consistent sensor streams and action representations, makes the dataset a challenging benchmark for multi-task policy learning in the wild.

We further analyze the real-world rollouts to better understand the self-correction behaviors exhibited by our model. 
As shown in Fig.~\ref{fig:real_rollout_app}, the robot occasionally fails to grasp or align the target object in early steps but subsequently re-evaluates the scene and retries to accomplish the task. 
Such behavior emerges despite being absent in training, suggesting that our method can help with temporal consistency and task persistence through the history-aware verification and feedback mechanism. 
This emergent robustness allows the agent to recover from minor action errors and maintain stable task completion in long-horizon scenarios.

\section{Limitation}

The latency limits the application in highly dynamic tasks, since our approach inherits the latency limitations of test-time scaling, which utilizes more inference computation to improve the performance. With more powerful hardware and asynchronous inference, the latency is expected to further decrease. Future work will explore lightweight solutions to mitigate the issue.

\section{Notation}
We summarize the key notations used throughout the proposed \textbf{E-TTS} framework in Table~\ref{tab:notations}. 
At each timestep $t$, the model observes the current scene $O_t$ and receives an instruction $I$ describing the target manipulation goal. 
Based on these inputs, the vision-language-action policy $\pi_{\theta}(c_t, a_t \mid O_t, I)$ jointly samples multiple reasoning–action pairs 
$J_t^{i,j} = (c_t^i, a_t^{i,j})$, where $c_t^i$ denotes the intermediate reasoning trace and $a_t^{i,j}$ the corresponding action candidate sample.

A history buffer $\mathcal{H}_t$ stores recent reasoning–action pairs and observations, providing temporal context for subsequent decision-making. 
To evaluate the sampled candidates, two verifiers are employed: the reasoning verifier $V_{\text{c}}$, which scores the semantic and logical soundness of reasoning outputs, and the action verifier $V_{\text{a}}$, which assesses the physical feasibility and task relevance of action predictions. 
Their outputs, $S_{\text{c}}^{i,j}$ and $S_{\text{a}}^{i,j}$, are normalized and combined into a joint score 
$S_t^{i,j} = S_{\text{c}}^{i,j} \times \hat{S}_{\text{a}}^{i,j}$ 
to identify the most coherent and executable reasoning–action pair.

During online inference, an $\epsilon$-greedy strategy is adopted to balance exploration and exploitation. 
When all candidates fall below a confidence threshold $\eta$, a feedback-guided refinement process is triggered: 
the reasoning verifier produces structured textual feedback $F_t^i$ explaining failure causes, which is concatenated with the original instruction $I$ to form an updated prompt $I_t^{i+1}$. 
This closed-loop design allows \textbf{E-TTS} to progressively improve its reasoning alignment and action precision through iterative refinement.

\end{document}

%% file: table/ecot.tex
\begin{table}[t]
  \caption{Performance comparison on three embodied manipulation tasks on Simpler widowx robot (Visual Matching). 
  The proposed method (E-CoT + \textbf{Ours}) consistently outperforms all baselines and ablation variants. 
  Gray rows denote ablation settings removing specific components such as feedback, scaling, or history buffer.}
  \label{tab:ecot}
  \centering
  \scalebox{0.7}{
  \begin{tabular}{@{}lcccc@{}}
    \toprule
    \textbf{Method} & \textbf{Spoon on Towel} & \textbf{Eggplant in Basket} & \textbf{Carrot on Plate} & \textbf{Average} \\
    \midrule
    E-CoT & 20.00 & 0.00 & 0.00 & 6.67 \\
    E-CoT + naive TTS & 41.67 & 4.17 & 25.00 & 23.61 \\
    E-CoT + Robomonkey & 33.33 & 8.33 & 37.50 &  26.38\\
    \rowcolor{gray!10} w/o feedback & 50.00 & 11.50 & 13.63 & 25.04 \\
    \rowcolor{gray!10} w/o reasoning scaling & 50.00 & 4.17 & 25.00 & 26.39 \\
    \rowcolor{gray!10} w/o action scaling & 45.83 & 16.67 & 29.17 & 30.56 \\
    \rowcolor{gray!10} w/o joint scoring & 42.85 & 18.75 & 31.80 & 31.13 \\
    \rowcolor{gray!10} w/o history buffer & 54.17 & 20.83 & 20.83 & 31.94 \\
    \rowcolor{gray!10} w/o $\epsilon$-greedy & 50.00 & 20.83 & 37.50 & 36.11 \\
    E-CoT + \textbf{Ours} & \textbf{58.33} & \textbf{22.22} & \textbf{38.89} & \textbf{39.81} \\
    \bottomrule
  \end{tabular}
  }
\end{table}

%% file: table/molmoact.tex
\begin{table*}[t]
\centering
\caption{Comparison of visual and variant performance across tasks on SimplerEnv. MolmoAct + \textbf{Ours}, significantly outperforms the MolmoAct and baselines, achieving the highest average scores in both Visual (80.00\%) and Variant (77.77\%) settings.}
\label{tab:molmoact_simpler}
\renewcommand{\arraystretch}{1.2}
  \scalebox{0.55}{
\begin{tabular}{@{}lcccccccc@{}}
\toprule
\multirow{2}{*}{\textbf{Method}} & \multicolumn{4}{c}{\textbf{Visual}} & \multicolumn{4}{c}{\textbf{Variant}} \\ 
\cmidrule(lr){2-5} \cmidrule(lr){6-9}
 & Pick Coke Can & Move Near & Open/Close Drawer & Average & Pick Coke Can & Move Near & Open/Close Drawer & Average \\
\midrule
MolmoAct & 74.50	&75.45&	63.25&	71.05	&66.95	&52.55&	77.75	&65.68\\
MolmoAct + naive TTS & 76.47 & 80.00 & 66.60 & 74.36 & 56.00 & 50.00 & 80.00 & 62.00 \\
MolmoAct + \textbf{Ours} & \textbf{83.00} & \textbf{85.00} & \textbf{72.00} & \textbf{80.00} & \textbf{80.00} & \textbf{70.00} & \textbf{83.30} & \textbf{77.77} \\
\bottomrule
\end{tabular}
}
\end{table*}

%% file: table/libero_molmoact.tex
\begin{table}[h]
\centering
\caption{\small Performance comparison on the \textbf{LIBERO} benchmark and \textbf{LIBERO-Plus} benchmark across four reasoning categories (Spatial, Object, Goal, Long-Horizon). 
MolmoAct + \textbf{Ours} consistently outperforms the baseline and naive variants.}
\label{tab:libero_side_by_side}
\begin{minipage}{0.48\linewidth}
\centering
\resizebox{0.85\linewidth}{!}{%
\begin{tabular}{lcccc}
\toprule
\textbf{Setting} & \textbf{Spatial} & \textbf{Goal} & \textbf{Object} & \textbf{Long} \\
\midrule
MolMoAct & 88.80 & 84.30 & 90.40 & 77.46 \\
MolMoAct + Ours & \textbf{91.20} & \textbf{85.04} & \textbf{91.36} & \textbf{80.99} \\
\bottomrule
\end{tabular}
}
\caption*{\small (a) LIBERO-Plus Benchmark}
\end{minipage}
\hfill
\begin{minipage}{0.48\linewidth}
\centering
\scalebox{0.55}{
\begin{tabular}{@{}lccccc@{}}
\toprule
\textbf{Method} & \textbf{Spatial} & \textbf{Object} & \textbf{Goal} & \textbf{Long} & \textbf{Average} \\
\midrule
MolmoAct & 87.00 & 92.67 & 87.60 & 75.00 & 85.57 \\
MolmoAct + naive TTS & 88.90 & 93.10 & 70.00 & 75.00 & 81.75 \\
MolmoAct + \textbf{Ours} & \textbf{93.60} & \textbf{97.50} & \textbf{92.00} & \textbf{80.00} & \textbf{90.78} \\
\bottomrule
\end{tabular}
}
\caption*{\small (b) LIBERO Benchmark}
\end{minipage}
\end{table}

%% file: table/rebuttal/vlabench.tex
\begin{table}[htbp]
\centering
\caption{\small Success Rate (SR) for representative tasks on the \textbf{VLABench} benchmark.}
\label{tab:vlabench_results_horizontal}

\begin{subtable}[t]{0.48\linewidth}
\centering
\resizebox{0.8\linewidth}{!}{%
\begin{tabular}{lccccccc}
\hline
\textbf{Track} & \textbf{IF} & \textbf{SB} & \textbf{SCT} & \textbf{SPo} & \textbf{AC} & \textbf{SD} & \textbf{Avg.} \\
\hline
Track 1 & 0.18 & 0.48 & 0.48 & 0.46 & 0.50 & 0.42 & 0.39 \\
Track 2 & -    & 0.02 & 0.15 & 0.40 & 0.06 & 0.14 & 0.20 \\
Track 3 & 0.08 & \textbf{0.38} & 0.50 & -    & 0.14 & 0.14    & 0.19 \\
Track 4 & 0.04 & 0.38 & 0.21 & 0.06 & 0.10 & 0.20 & 0.15 \\
Track 6 & \textbf{0.06} & 0.45 & 0.29 & 0.22 & \textbf{0.46} & 0.30 & 0.25 \\
\hline
\end{tabular}%
}
\subcaption*{(a) $\pi_{0.5}$}
\end{subtable}
\hfill
\begin{subtable}[t]{0.48\linewidth}
\centering
\resizebox{0.8\linewidth}{!}{%
\begin{tabular}{lccccccc}
\hline
\textbf{Track} & \textbf{IF} & \textbf{SB} & \textbf{SCT} & \textbf{SPo} & \textbf{AC} & \textbf{SD} & \textbf{Avg.} \\
\hline
Track 1 & \textbf{0.20} & \textbf{0.53} & \textbf{0.53} & \textbf{0.63} & \textbf{0.54} & \textbf{0.48} & \textbf{0.42} \\
Track 2 & - & \textbf{0.02} & \textbf{0.16} & \textbf{0.48} & \textbf{0.08} & \textbf{0.14} & \textbf{0.24} \\
Track 3 & \textbf{0.10} & 0.36 & \textbf{0.54} & - & \textbf{0.18} & \textbf{0.16} & \textbf{0.21} \\
Track 4 & \textbf{0.06} & \textbf{0.48} & \textbf{0.24} & \textbf{0.08} & \textbf{0.16} & \textbf{0.22} & \textbf{0.16} \\
Track 6 & 0.04 & \textbf{0.48} & \textbf{0.31} & \textbf{0.28} & \textbf{0.31} & \textbf{0.34} & \textbf{0.26} \\
\hline
\end{tabular}%
}
\subcaption*{(b) $\pi_{0.5}$+Ours}
\end{subtable}
\end{table}

%% file: table/rebuttal/Q2.tex

\begin{table}[htbp]
\centering
\caption{\small Ablations on \textbf{SimplerEnv}. R/A is reasoning/action.
}
\label{tab:latency_ablation}

\begin{subtable}[t]{0.49\linewidth}
\centering
\resizebox{0.8\linewidth}{!}{
\begin{tabular}{lccc}
\toprule
Method & R/A Samples & SR (\%) & Avg Time (s) \\
\midrule
E-CoT  & -- & 20.00 & 8.35 \\
E-CoT + ours & 10 / 20 & 37.50 & 14.57 \\
E-CoT + ours & 10 / 30 & 41.67 & 11.46 \\
E-CoT + ours & 10 / 50 & 50.00 & 12.23 \\
E-CoT + ours & 20 / 20 & 59.10 & 14.77 \\
E-CoT + ours & 50 / 100 & 50.00 & 15.65 \\
E-CoT + ours & 100 / 100 & 62.50 & 20.22 \\
\bottomrule
\end{tabular}
}
\subcaption*{(a) Ablations on sample scale and efficiency}
\end{subtable}
\hfill
\begin{subtable}[t]{0.49\linewidth}
\centering
\resizebox{0.7\linewidth}{!}{
\begin{tabular}{c c c c c c}
\toprule
$K$ & SR (\%) & $\eta$ & SR(\%) & $\epsilon$ & SR(\%) \\
\midrule
1  & 37.50 & 0.1 & \textbf{70.83} & 0.1 & \textbf{54.17} \\
3  & 33.33 & 0.2 & 45.83           & 0.2 & 50.00           \\
5  & 45.83 & 0.3 & 50.00           & 0.3 & 37.50           \\
10 & \textbf{58.33} & 0.4 & 37.50     & 0.5 & 37.50           \\
20 & 45.83 & 0.5 & 33.33           & 0.7 & 33.33           \\
\bottomrule
\end{tabular}
}
\subcaption*{(b) Ablations on hyperparameter}
\end{subtable}

\end{table}

%% file: table/algorithm.tex
\begin{algorithm}[htbp]
\caption{E-TTS Pseudocode}
\label{alg:joint-inference}
\textbf{Input:} policy $\pi_{\theta}$, reasoning verifier $V_{\text{c}}$, action verifier $V_{\text{a}}$, feedback advisor $f$, instruction $I$, initial observation $O$, history length $K$, max reasoning samples $M$, max action samples $N$, selection parameter $\epsilon$, acceptance threshold $\eta$, max timesteps $T$, evaluation steps $\Delta$, the set of the best sample in each batch $\mathcal{A}$.

\begin{algorithmic}[1]
    \STATE Initialize $\mathcal{H}_0 = \emptyset$, $t \leftarrow 0$
    \FOR{$t = \Delta$ to $T$ step $\Delta$}
        \STATE Obtain observation $O_t$ and initialize $\mathcal{A} \leftarrow \emptyset$
        \FOR{$i = 1$ to $M$} 
            \STATE Sample reasoning $c_t^i \sim p_\theta(c \mid O_t, I, F_{t}^{i-1})$
            \STATE Form joint pairs $J_t^{i,j} = (c_t^i, a_t^{i,j})$ with $a_t^{i,j} \sim p_\theta(a \mid c_t^i, O_t, I), \; j = 1\dots N$
            \STATE Build joint batch $\mathcal{B}_t^i = \{ J_t^{i,j} \}_{j=1}^{N}$
            \STATE Build history buffer: $\mathcal{H}_t = \{ J_{t-K}, \dots, J_{t-1} \}$
        \FOR{each $J_t^{i,j} \in \mathcal{B}_t^i$}
            \STATE $S_{c}^{i,j} \leftarrow V_{\text{c}}(\mathcal{H}_t, J_t^{i,j}, O_t, I)$
            \STATE $S_{a}^{i,j} \leftarrow V_{\text{a}}(J_t^{i,j}, O_t, I)$
            \STATE $S_t^{i,j} \leftarrow S_c^{i,j} \times \hat{S}_a^{i,j}$
        \ENDFOR
        \STATE Sample $u \sim \text{Uniform}(0,1)$
        \STATE $j^* \leftarrow \arg\max_j S_t^{i,j}$
        \IF{$u \le \epsilon$} 
            \STATE Execute $a_t$ from $\text{Random}(\mathcal{B}_t^i)$
        \ELSIF{$S_t^{i,j^*} \le \eta$}
            \STATE Generate feedback $F_{t}^{i} \leftarrow f(\mathcal{H}_t, J_t^{i,j^*}, O_t, I)$
            \STATE Update $I \leftarrow \operatorname{Concat}(I, F_{t}^{i})$
            \STATE Update $\mathcal{A} \leftarrow J_t^{i,j^*}$
        \ELSE
            \STATE Execute $a_t = J_t^{i,j^*}$
        \ENDIF
        \ENDFOR
        \IF{\text{no sample executed}}
            \STATE $i^{*} \leftarrow \arg\max_i \mathcal{A}$
            \STATE Execute $a_t = J_t^{i^*,j^*}$
        \ENDIF
    \ENDFOR
\end{algorithmic}
\end{algorithm}

%% file: table/appendix_feedback_case.tex
\begin{table*}[htbp]
\centering
\caption{Representative batches showing CoT scores, action rewards, joint scores, and two-line feedback summaries.}
\small
\resizebox{0.8\linewidth}{!}{
\begin{tabular}{c|c|c|c|c}
\toprule
Step-Batch & CoT Score & Max Action Reward & Joint Score & Supervisor Feedback \\
\midrule
10-1 & 0.1754 & 0.4385 & 0.0769 & Robot moved towards spoon but selected incorrect one;\\ & & & & should distinguish correct spoon relative to towel. \\
10-2 & 0.0608 & 0.2494 & 0.0152 & Robot moved forward without aligning gripper;\\ & & & &  feedback suggests adjusting gripper position to improve grasp. \\
10-3 & 0.1296 & 0.3972 & 0.0515 & Robot still misaligned; feedback reinforces correct gripper\\ & & & &  alignment and positioning. \\
10-4 & 0.5883 & 0.3403 & 0.2002 & - \\
\bottomrule
\end{tabular}
}
\label{tab:fgir_feedback}
\end{table*}

%% file: table/app_robo_ours.tex
\begin{table}[htbp]
\centering
\caption{Comparison of success rate (\%) and inference latency (s/step) on Simpler Env. Our method achieves a significant performance gain with marginal computational overhead.}
\label{tab:performance_comparison_robo_ours}
\resizebox{0.85\linewidth}{!}{
\begin{tabular}{lcccccccc}
\toprule
\multirow{2}{*}{Method} & \multicolumn{2}{c}{Spoon on Towel} & \multicolumn{2}{c}{Eggplant in Basket} & \multicolumn{2}{c}{Carrot on Plate} & \multicolumn{2}{c}{Average} \\
\cmidrule(lr){2-3} \cmidrule(lr){4-5} \cmidrule(lr){6-7} \cmidrule(lr){8-9}
 & SR$\uparrow$ & Time$\downarrow$ & SR$\uparrow$ & Time$\downarrow$ & SR$\uparrow$ & Time$\downarrow$ & SR$\uparrow$ & Time$\downarrow$ \\
\midrule
E-CoT + Robomonkey & 33.33 & 12.81 & 8.33 & 11.75 & 37.50 & 12.81 & 26.38 & 12.46 \\
E-CoT + \textbf{Ours} & \textbf{59.10} & 14.77 & \textbf{22.73} & 13.02 & \textbf{38.10} & 12.49 & \textbf{39.98} & 13.43 \\
\bottomrule
\end{tabular}
}
\end{table}


%% file: table/appendix_er1.tex
\begin{table*}[t]
\centering
\caption{Quantitative comparison on the Simpler WidowX benchmark. 
E-TTS (ER-1+Ours) achieves consistent performance gains across all tasks. 
Success rate is reported as the percentage of successful executions.}
\label{tab:simpler_result_append_er1}
\resizebox{0.8\linewidth}{!}{
\begin{tabular}{lcccc}
\toprule
Method & Spoon on Towel & Eggplant in Basket & Carrot on Plate & Average \\
\midrule
ER-1 & 62.50\% & 4.17\% & 52.78\% & 39.82\% \\
ER-1 + naive TTS & 63.88\% & 5.55\% & 51.38\% & 40.27\% \\
\textbf{ER-1 + Ours (E-TTS)} & \textbf{65.27\%} & \textbf{8.33\%} & \textbf{61.11\%} & \textbf{44.90\%} \\
\bottomrule
\end{tabular}
}
\end{table*}

%% file: table/app_vlabench.tex
\begin{table*}[htbp]
\centering

\caption{\small Performance comparison on the \textbf{VLaBench} benchmark. We report the \textbf{Success Rate (SR)} for various tasks.}
\label{tab:vlabench_results}

\footnotesize
\resizebox{0.85\textwidth}{!}{
\begin{tabular}{llccccccccccccc}
\toprule
\textbf{Method} & \textbf{Track} & \textbf{AC} & \textbf{IF} & \textbf{SB} & \textbf{SCT} & \textbf{SD} & \textbf{SF} & \textbf{SM} & \textbf{SNLP} & \textbf{SPa} & \textbf{SPo} & \textbf{ST} & \textbf{SUM} & \textbf{Avg.} \\
\midrule
$\pi_{0.5}$ & Track 1 & 0.50 & 0.18 & 0.48 & 0.48 & 0.42 & 0.38 & \textbf{0.44} & -- & 0.28 & 0.46 & 0.28 & -- & 0.39 \\
      & Track 2 & 0.06 & -- & 0.02 & 0.15 & 0.14 & \textbf{0.44} & 0.20 & -- & 0.28 & 0.40 & 0.14 & -- & 0.20 \\
      & Track 3 & 0.14 & 0.08 & \textbf{0.38} & 0.50 & 0.14 & 0.24 & -- & 0.14 & 0.16 & -- & 0.04 & 0.04 & 0.19 \\
      & Track 4 & 0.10 & 0.04 & 0.38 & 0.21 & 0.20 & \textbf{0.18} & 0.10 & -- & \textbf{0.12} & 0.06 & \textbf{0.08} & -- & 0.15 \\
      & Track 6 & \textbf{0.46} & \textbf{0.06} & 0.45 & 0.29 & 0.30 & 0.24 & \textbf{0.13} & -- & 0.28 & 0.22 & \textbf{0.04} & -- & 0.25 \\
\midrule
$\pi_{0.5}$+Ours & Track 1 & \textbf{0.54} & \textbf{0.20} & \textbf{0.53} & \textbf{0.53} & \textbf{0.48} & \textbf{0.40} & 0.33 & -- & \textbf{0.30} & \textbf{0.63} & \textbf{0.32} & -- & \textbf{0.42} \\
          & Track 2 & \textbf{0.08} & -- & \textbf{0.02} & \textbf{0.16} & \textbf{0.14} & 0.38 & \textbf{0.33} & -- & \textbf{0.30} & \textbf{0.48} & \textbf{0.22} & -- & \textbf{0.24} \\
          & Track 3 & \textbf{0.18} & \textbf{0.10} & 0.36 & \textbf{0.54} & \textbf{0.16} & \textbf{0.24} & -- & \textbf{0.16} & \textbf{0.26} & -- & \textbf{0.06} & \textbf{0.09} & \textbf{0.21} \\
          & Track 4 & \textbf{0.16} & \textbf{0.06} & \textbf{0.48} & \textbf{0.24} & \textbf{0.22} & 0.14 & \textbf{0.14} & -- & 0.10 & \textbf{0.08} & 0.06 & -- & \textbf{0.16} \\
          & Track 6 & 0.43 & 0.04 & \textbf{0.48} & \textbf{0.31} & \textbf{0.34} & \textbf{0.26} & 0.11 & -- & \textbf{0.32} & \textbf{0.28} & 0.03 & -- & \textbf{0.26} \\
\bottomrule
\multicolumn{15}{p{\textwidth}}{\scriptsize \textbf{Note:} \textbf{SR}: Success Rate. Abbreviations: \textbf{AC}: add\_condiment, \textbf{IF}: insert\_flower, \textbf{SB}: select\_book, \textbf{SCT}: select\_chemistry\_tube, \textbf{SD}: select\_drink, \textbf{SF}: select\_fruit, \textbf{SM}: select\_mahjong, \textbf{SNLP}: select\_nth\_largest\_poker, \textbf{SPa}: select\_painting, \textbf{SPo}: select\_poker, \textbf{ST}: select\_toy, \textbf{SUM}: select\_unique\_type\_mahjong.}
\end{tabular}
}
\end{table*}

%% file: table/app_real_data.tex
\begin{table*}[t]
    \centering
    \caption{Statistics of the real-world dataset.}
    \label{tab:dataset}
    \begin{tabular}{lccc}
        \toprule
        Task description & Episodes & Frames & Avg.\ length \\
        \midrule
        Pick up the toy lion and place it on the shelf & 100 & 4{,}577 & 45.8 \\
        Hang the bag on the shelf & 100 & 2{,}917 & 29.2 \\
        Press the sanitizer & 100 & 1{,}385 & 13.8 \\
        Put the Coke can on the upper drawer & 100 & 8{,}899 & 89.0 \\
        \bottomrule
    \end{tabular}
\end{table*}

%% file: table/notation.tex
\begin{table*}[t]
\centering
\caption{\textbf{List of Notations.} Summary of the mathematical symbols used throughout the E-TTS framework.}
\label{tab:notations}
\resizebox{0.85\linewidth}{!}{
\begin{tabular}{ll}
\toprule
\textbf{Notation} & \textbf{Description} \\
\midrule
$I$ & Task instruction describing the embodied manipulation goal. \\
$O_t$ & Observation (visual input) at timestep $t$. \\
$c_t$ & Reasoning output (intermediate cognitive state) at timestep $t$. \\
$c_t^i$ & $i$-th reasoning output (intermediate cognitive state) at timestep $t$. \\
$a_t$ & Action predicted by the policy at timestep $t$. \\
$a_t^{i,j}$ & $j$-th action predicted by the policy with $c_t^i$ at timestep $t$. \\
$J_t^{i,j} = (c_t^i, a_t^{i,j})$ & Joint reasoning–action pair sampled at timestep $t$. \\
$\mathcal{B}_t^i = \{ J_t^{i,j} \}_{j=1}^{N}$ & Joint batch for the $i$-th joint reasoning–action pair. \\
$\pi_{\theta}(c_t, a_t \mid O_t, I)$ & Vision-Language-Action policy parameterized by $\theta$. \\
$M$ & Max Number of reasoning samples. \\
$N$ & Number of action samples per reasoning trace. \\
$\mathcal{H}_t = \{ J_{t-K}, O_{t-K}, \ldots, J_{t-1}, O_{t-1} \}$ & History buffer storing past reasoning–action pairs and observations. \\
$K$ & Window size (length) of the history buffer. \\
$V_{\text{c}}$ & Reasoning verifier evaluating reasoning quality and consistency. \\
$V_{\text{a}}$ & Action verifier assessing action feasibility and relevance. \\
$S_{\text{c}}^{i,j}$ & Confidence score from the reasoning verifier. \\
$S_{\text{a}}^{i,j}$ & Confidence score from the action verifier. \\
$\hat{S}_{\text{a}}^{i,j}$ & Normalized action score. \\
$S_t^{i,j} = S_{\text{c}}^{i,j} \times \hat{S}_{\text{a}}^{i,j}$ & Final joint score combining reasoning and action. \\
$\epsilon$ & Exploration rate in the $\epsilon$-greedy selection policy. \\
$\eta$ & Confidence threshold for accepting joint samples. \\
$u \sim \text{Uniform}(0,1)$ & Random variable controlling exploration vs. exploitation. \\
$F_t^i$ & Feedback text describing reasoning flaws and corrective guidance. \\
$I_t^{i+1} = \operatorname{Concat}(I, F_t^i)$ & Refined instruction after feedback-guided update. \\
\bottomrule
\end{tabular}
}
\end{table*}

%% file: table/taco.tex
\begin{table}[htbp]
\centering
\caption{Success rates (\%) on LIBERO-Long tasks.}
\label{tab:taco}
\resizebox{0.95\linewidth}{!}{
\begin{tabular}{lccccccccccc}
\toprule
Method & 
Soup/Sauce & 
Chs/Btr & 
Stove/Moka & 
Blk Bowl & 
Mug/Plate & 
Bk/Caddy & 
Mug/Pud & 
Soup/Chs & 
Moka/Stv & 
Mug/Micro & 
Avg. \\
\midrule
$\pi_{0.5}$  & \textbf{98.0} & \textbf{100.0} & 98.0 & 98.0 & \textbf{98.0} & \textbf{100.0} & \textbf{96.0} & 94.0 & 68.0 & \textbf{98.0} & 94.8 \\
$\pi_{0.5}$ + TACO               & 96.0 & 98.0 & \textbf{100.0} & 98.0 & 96.0 & \textbf{100.0} & \textbf{96.0} & \textbf{100.0} & 76.0 & \textbf{98.0} & 95.8 \\
$\pi_{0.5}$ + Ours               & 96.0 & 98.0 & \textbf{100.0} & \textbf{100.0} & \textbf{98.0} & \textbf{100.0} & \textbf{96.0} & 98.0 & \textbf{80.0} & 96.0 & \textbf{96.2} \\
\bottomrule
\end{tabular}
}
\end{table}